\documentclass[11pt]{article}

\usepackage[margin=1in]{geometry}
\usepackage{authblk}
\usepackage{graphicx}
\usepackage{amsmath, amssymb}
\usepackage{hyperref}
\usepackage{subcaption}
\usepackage[numbers,sort&compress]{natbib} 
\usepackage{times}
\usepackage{setspace}
\usepackage[font=small,labelfont=bf]{caption} 
\usepackage{lineno}
\usepackage{multirow}
\usepackage{helvet}
\usepackage{courier}
\usepackage{algorithm}
\usepackage{algorithmic}
\usepackage{tabularx}
\usepackage{booktabs, color}
\usepackage{comment}
\title{\textbf{Improving Hospital Risk Prediction with Knowledge-Augmented Multimodal
EHR Modeling}}

\author{
    Rituparna Datta\textsuperscript{\rm 1},
    Jiaming Cui\textsuperscript{\rm 3},
    Zihan Guan\textsuperscript{\rm 1},
    Vishal G. Reddy\textsuperscript{\rm 2},
    Joshua C. Eby\textsuperscript{\rm 5},
    Gregory Madden\textsuperscript{\rm 5},
    Rupesh Silwal\textsuperscript{\rm 5},\\
    Anil Vullikanti\textsuperscript{\rm 1,4}
}
\affil{
    \textsuperscript{\rm 1} Department of Computer Science, University of Virginia\\
    \textsuperscript{\rm 2} University of Virginia School of Medicine\\
    \textsuperscript{\rm 3}Virginia Polytechnic Institute and State University\\
    \textsuperscript{\rm 4} Biocomplexity Institute and Initiative, University of Virginia\\
    \textsuperscript{\rm 5}Division of Infectious Diseases \& International Health, University of Virginia School of Medicine
}
\author{}
\date{} 
\newcommand{\tool}{KAMELEON}
\newcommand{\anil}[1]{\textcolor{blue}{#1}}

\newcommand{\zeroshot}[1]{\textit{#1}$^{\dagger}$}

\begin{document}

\maketitle
\begin{abstract}
Accurate prediction of clinical outcomes using Electronic Health Records (EHRs) is critical for early intervention, efficient resource allocation, and improved patient care. 
EHRs contain multimodal data, including both structured data and unstructured clinical notes that provide rich, context-specific information. In this work, we introduce a unified framework that seamlessly integrates these diverse modalities, leveraging all relevant available information through a two-stage architecture for clinical risk prediction. In the first stage, a fine-tuned Large Language Model (LLM) extracts crucial, task-relevant information from clinical notes, which is enhanced by graph-based retrieval of external domain knowledge from sources such as a medical corpus like PubMed, grounding the LLM's understanding. The second stage combines both unstructured representations and features derived from the structured data to generate the final predictions. 
This approach supports a wide range of clinical tasks. Here, we demonstrate its effectiveness on 30-day readmission and in-hospital mortality prediction. Experimental results show that our framework achieves strong performance, with AUC scores of $0.84$ and $0.92$, respectively, despite these tasks involving severely imbalanced datasets, with positive rates ranging from approximately $4\%$ to $13\%$. 
Moreover, it outperforms all existing baselines and clinical practices, including established risk scoring systems. 
To the best of our knowledge, this is one of the first frameworks for healthcare prediction which enhances the power of an LLM-based graph-guided knowledge retrieval method by combining it with structured data for improved clinical outcome prediction.
\end{abstract}
\section{Introduction}

Appropriate use of clinical prediction tools for early identification of high-risk patients for different conditions allows for clinical decision making, timely interventions, escalation of care, intensive monitoring, and identification of gaps in outpatient management, e.g.,~\cite{cai2016real, kong2020using, brajer2020prospective, kansagara2011risk, mahmoudi2020use}. For instance, readmission within a short period is a priority under many regulatory frameworks and value-based care models, where high readmission rates may lead to financial penalties, e.g.,~\cite{upadhyay2019readmission, clement2017will}.
Therefore, effective models of short-term risk prediction (e.g., 30 days) can guide targeted interventions, including more detailed discharge instructions, closer post-discharge monitoring, or referrals to transitional care programs. 
While traditional prediction tools relied on simple statistical models, e.g., regression and decision trees~\cite{brisimi2018predicting, herazo2021decision}, for risk assessment, more complex machine learning methods are increasingly applied to clinical prediction tasks, e.g.,~\cite{cui2025identifying,arsalan2025enhancing,yu2024smart,wu2023medlink,zhu2024emerge}.
In this work, we focus on two commonly studied clinical problems: 
(1) 30-day readmission prediction, which determines whether a patient will be readmitted to the hospital within 30 days after discharge, and (2) mortality prediction, which determines the patient’s in-hospital mortality status.

\begin{figure*}[h]
    \centering   \includegraphics[width=\linewidth]{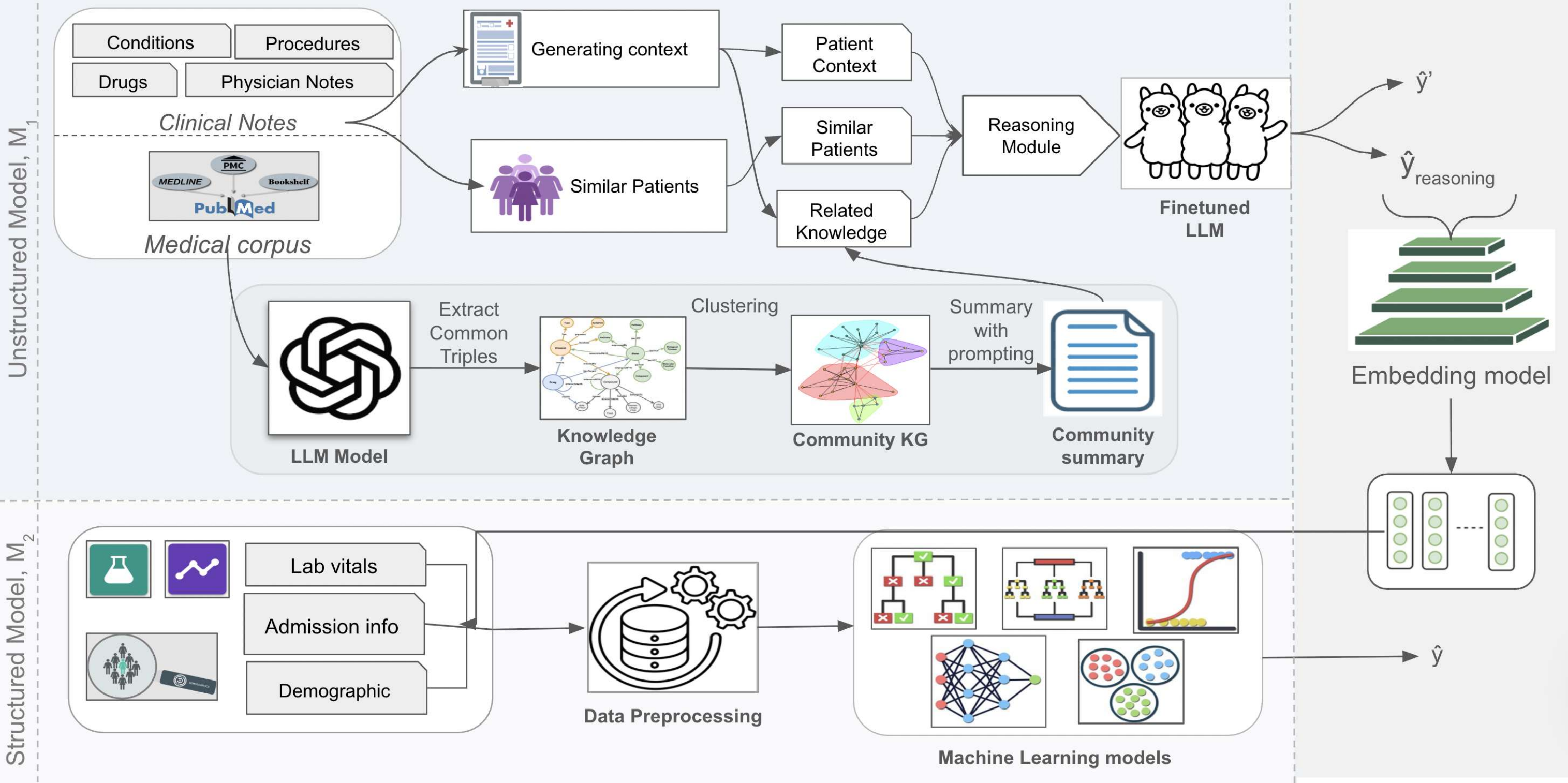}
    \caption{A two-stage hybrid framework for predictive tasks, integrating structured and unstructured patient data with Large Language Models (LLMs). \textbf{Step 1 ($M_1$)} focuses on knowledge-enhanced context generation with an initial LLM output, while \textbf{Step 2 ($M_2$)} integrates the finetuned LLM outputs with structured data by creating an embedding, for final ML prediction. }
    \label{fig:framework}
\end{figure*}

There has been a lot of work on developing diverse kinds of machine learning methods for these problems using 
Electronic Health Record (EHR) data, which contain rich information on patient health, e.g.,~\cite{cai2016real, jiangreasoning,qiu2025deep,yang2023transformehr, brisimi2018predicting, bellot2019bayesian, cui2025identifying,arsalan2025enhancing,yu2024smart,wu2023medlink,zhu2024emerge}. 
Most of this work focuses on structured EHR data, which spans various types and representations~\cite{hartvigsen2018early}, including admission/discharge information, procedures and interventions, medications, lab orders and results, billing codes (e.g., ICD, CPT), physiological time-series (e.g., vital signs), imaging data, and clinical documentation such as progress notes or discharge summaries. 
Unstructured data from clinical notes has been used in a fairly simple manner, such as bag of words, TF-IDF \cite{multimodal, HUANG2019141} to facilitate the use of conventional machine learning methods for clinical tasks. 
Clinical notes are complex and poorly structured, which limits their utility in clinical informatics tasks, even when using advanced natural language processing (NLP) techniques. 
While large language models (LLMs) offer a powerful means to process such notes, especially when combined with large biomedical datasets to capture richer semantics beyond keywords and embeddings~\cite{shi2024ehragent, li2024scoping, singhal2022large, jiangreasoning}, they still face significant limitations, such as  
hallucinations, factual inaccuracies, and inadequate domain grounding~\cite{shi2024ehragent, jiangreasoning}. 
For instance, models such as Med-PaLM~\cite{singhal2022large} exhibit strong language generation capabilities but frequently misinterpret similar-sounding medical terms. 
Recent approaches have attempted to enhance LLMs with structured knowledge via graph-based retrieval (e.g., GraphRAG), but their performance remains limited due to a lack of explicit reasoning, e.g.,~\cite{ye2021medretriever, jiang2024graphcare, xu2024ramehr, niu2024ehr, zhu2024emerge}.
In recent work, Jiang et al.~\cite{jiangreasoning} developed KARE, a graphRAG and context augmentation approach, for clinical prediction tasks, which address many challenges associated with using LLMs for clinical tasks on the MIMIC dataset.

However, the performance of all prior methods remains limited because
clinical tasks using EHRs present many non-trivial challenges:
(1) Multi-modality of clinical data: The presence of both structured data and unstructured text requires methods capable of effectively learning from both modalities. (2) Long-context textual data: Clinical text often contains a mix of specialized medical terminology and informal or colloquial expressions, making information retrieval challenging. (3) Severe class imbalance: Prediction tasks are typically highly imbalanced. For example, only about 4\% of patients are readmitted within 30 days, resulting in heavily skewed training data.

Here, we develop a novel framework, \textbf{\tool{}} (\textbf{K}nowledge-\textbf{A}ugmented \textbf{M}ultimodal \textbf{E}HR \textbf{LE}arning for \textbf{O}utcome predictio\textbf{N})
that addresses the limitations of prior work by integrating multimodal EHR data (including structured clinical components and unstructured physician notes), and external biomedical knowledge.
\tool{} consists of two components (as shown in Figure \ref{fig:framework}):\\
(1) An unstructured model \( M_1 \) (Section ~\ref{sec:unstruct}) that processes clinical notes and retrieves relevant biomedical knowledge using a PubMed-derived graph and knowledge-augmented reasoning, and outputs a prediction for a patient, along with its reasoning; this extends the approach of~\cite{jiangreasoning}.
Physician notes in EHRs can be lengthy and exceed the context window, and these are summarized using an LLM, and used as context.
To introduce domain-level medical knowledge, we build a biomedical knowledge graph (KG) by combining the Unified Medical Language System (UMLS)\cite{bodenreider2004unified}, PubMed abstracts, and large language model (LLM)-generated entity-relation triples.
KG is partitioned into semantically coherent and well-connected clusters, and textual summary generated by an LLM for the most relevant clusters for each patient cluster, are used to enrich the context.
Furthermore, labeled context is added by identifying semantically similar patient visits, which is used to fine-tune the LLM.
Finally, $M_1$ produces a prediction for the patient, along with a reasoning.\\
(2) A structured model $\mathbf{M}_2$ (Section~\ref{sec:struct}), which extracts structured features from the patient's EHR for the stay, including 
(a) Static demographic and admission data,
(b) time-varying vitals (which are normalized and summarized, when used as features), and 
(c) diagnoses, procedures and medications.
In addition, $M_2$ includes the LLM’s prediction and its tokenized reasoning transformed into an embedding, as inputs.
Finally, different kinds of standard machine learning methods are trained in $M_2$ using these inputs. 
We first train $M_1$ separately, and use the LLM outputs to train $M_2$.

We demonstrate the effectiveness of \tool{} for the 30-day readmission risk and mortality prediction tasks, which have been studied extensively, both using MIMIC-III and other private EHRs from specific hospitals. We compare the performance with a number of structured ML and LLM baselines, with respect to multiple metrics (discussed in Section~\ref{sec:results}). \textbf{\emph{\tool{} consistently outperforms all prior work on MIMIC-III datasets}}. It also shows clear gains over the strongest unstructured LLM baseline (Llama3-Med42-8B). The only other prior work which has similar performance for 30-day readmission~\cite{jamiaopen} is on a Norwegian EHR dataset, which is significantly less imbalanced ($16.7\%$ readmission positive rate, instead of $4\%$ in the case of MIMIC-III).


In summary, \tool{} is the first systematic framework to enhance the power of LLMs for healthcare prediction tasks through graph-guided knowledge retrieval combined with structured machine learning methods. 
We expect this framework to be readily applicable to other clinical questions beyond those examined in this study.

\newcommand{\istrue}{\checkmark}
\newcommand{\isfalse}{\textendash} 
\begin{table*}[htbp]
\centering
\caption{Comparison of Models for 30-Day Readmission and In-Hospital Mortality Prediction. Scores from \textit{KARE}~\cite{jiangreasoning} are re-evaluated using our pipeline due to LLM differences and incorrect data preprocessing in their code. Reported metrics include Accuracy, Negative Predictive Value (NPV), Precision (Positive Predictive Value), Sensitivity (Recall), Specificity, Macro F1, AUROC, and AUPRC.
Models marked with $^{\dagger}$ are evaluated in a zero-shot setting without fine-tuning.
}
\label{tab:merged_model_tasks}
\resizebox{\textwidth}{!}{%
\begin{tabular}{l|p{1cm}|p{1cm}| p{1cm}p{1cm}p{1.5cm}p{1.5cm}p{1.5cm}p{1cm}p{1.2cm}p{1cm}} 
\toprule
\textbf{Model} & \multicolumn{1}{c|}{$\boldsymbol{\mathcal{D}^{\mathrm{struct}}}
$} & \multicolumn{1}{c|}{$\boldsymbol{\mathcal{D}^{\mathrm{unstruct}}}
$} & \textbf{Acc} & \textbf{NPV} & \textbf{Precision (PPV)} & \textbf{Specificity} & \textbf{Sensitivity} & \textbf{Macro F1} & \textbf{AUROC} & \textbf{AUPRC} \\
\midrule
\multicolumn{11}{c}{\textbf{Task: 30-Day Readmission Prediction}} \\
\midrule
LogReg (C=1.0) \cite{harutyunyan2019multitask} & \istrue & \isfalse & 0.831 & 0.869 & 0.036 & 0.951 & 0.013 & 0.464 & 0.463 & 0.090 \\
MLP \cite{jamiaopen}   & \istrue & \isfalse & 0.828 & 0.876 & 0.182 & 0.934 & 0.100 & 0.516 & 0.559 & 0.165 \\
BalancedRF     & \istrue & \isfalse & 0.760 & 0.970 & 0.070 & 0.780 & 0.430 & 0.490 & 0.673 & 0.066 \\
LSTM \cite{harutyunyan2019multitask}   & \istrue & \isfalse & 0.820 & 0.876 & 0.163 & 0.925 & 0.100 & 0.512 & 0.569 & 0.152 \\
\hline
\zeroshot{Claude-3.7-Sonnet} \cite{claude2024} & \isfalse & \istrue & 0.240 & 0.790 & 0.199 & 0.068 & 0.927 & 0.227 & 0.498 & 0.199 \\
\zeroshot{Medgemma-4b-it} \cite{medgemma2024}    & \isfalse & \istrue & 0.350 & 0.770 & 0.190 & 0.270 & 0.690 & 0.350 & 0.480 & 0.190 \\
\zeroshot{Llama3-Med42-8B} \cite{llama3med2024}  & \isfalse & \istrue & 0.390 & 0.800 & 0.210 & 0.360 & 0.670 & 0.410 & 0.510 & \textbf{0.210} \\
$M_1$ & \isfalse & \istrue & 0.660 & 0.870 & 0.130 & 0.720 & 0.280 & 0.480 & 0.506 & 0.195 \\
\textit{KARE} \cite{jiangreasoning} & \isfalse & \istrue & 0.271 & 0.785 & 0.191 & 0.131 & 0.851 & 0.269& 0.491 & 0.191 \\
\hline
\tool{}-{LogReg}, (C=1.0) & \istrue & \istrue & 0.833 & 0.869 & 0.037 & 0.953 & 0.013 & 0.519 & 0.130 & 0.148 \\
\tool{}-{LogReg}, (C=0.01) & \istrue & \istrue & 0.871 & 0.874 & \textbf{0.333} & \textbf{0.996} & 0.013 & 0.478 & 0.551 & 0.152 \\
\tool{}-{LSTM}  & \istrue & \istrue & 0.840 & 0.880 & 0.190 & 0.950 & 0.090 & 0.510 & 0.505 & 0.135 \\
\tool{}-{BalancedRF}     & \istrue & \istrue & 0.800 & \textbf{0.990} & 0.130 & 0.800 & \textbf{0.790} & 0.550 & \textbf{0.845} & 0.150 \\
\tool{}-{MLP} & \istrue & \istrue & \textbf{0.910} & 0.970 & 0.160 & 0.940 & 0.280 & \textbf{0.580} & 0.820 & 0.138 \\
\midrule
\multicolumn{11}{c}{\textbf{Task: In-Hospital Mortality Prediction}} \\
\midrule
Logistic Regression \cite{harutyunyan2019multitask}& \istrue & \isfalse & 0.850 & 0.912 & 0.340 & 0.916 & 0.331 & 0.625 & 0.624 & 0.190 \\
LSTM  \cite{harutyunyan2019multitask}  & \istrue & \isfalse & 0.690 & 0.800 & 0.260 & 0.800 & 0.250 & 0.530 & 0.560 & 0.240 \\
BRF     & \istrue & \isfalse & 0.810 & 0.950 & 0.340 & 0.820 & 0.700 & 0.670 & 0.860 & 0.475 \\
LightGBM \cite{jamiaopen} & \istrue & \isfalse & 0.890 & 0.930 & 0.510 & 0.940 & 0.480 & 0.720 & 0.866 & 0.534 \\
MLP \cite{jamiaopen}  & \istrue & \isfalse & 0.870 & 0.920 & 0.430 & 0.920 & 0.430 & 0.680 & 0.829 & 0.426 \\
XGBoost & \istrue & \isfalse & 0.890 & 0.920 & 0.520 & 0.950 & 0.380 & 0.695 & 0.835 & 0.487 \\
\hline
\zeroshot{Claude-3.7-Sonnet} \cite{claude2024} & \isfalse & \istrue & 0.800 & 0.890 & 0.120 & 0.880 & 0.130 & 0.510 & 0.510 & 0.110 \\

\zeroshot{Medgemma-4b-it} \cite{medgemma2024}   & \isfalse & \istrue & 0.120 & 0.950 & 0.100 & 0.020 & 0.990 & 0.120 & 0.510 & 0.110 \\

\zeroshot{Llama3-Med42-8B} \cite{llama3med2024}   & \isfalse & \istrue & 0.160 & 0.950 & 0.120 & 0.100 & 0.970 & 0.190 & 0.530 & 0.120 \\
\textit{$M_1$} &\isfalse & \istrue& 0.614 & 0.890 & 0.134 & 0.641 & 0.413 & 0.474 & 0.527&0.125\\
\textit{KARE} \cite{jiangreasoning} & \isfalse & \istrue & 0.639 & 0.885 & 0.129 & 0.678 & 0.353 & 0.478 & 0.515 & 0.122\\
\hline
\tool{}-{BRF}     & \istrue & \istrue & 0.880 & 0.930 & 0.490 & 0.934 & 0.492 & 0.710 & 0.876 & 0.543 \\
\tool{}-{LSTM}   & \istrue & \istrue & 0.730 & 0.820 & 0.430 & 0.840 & 0.390 & 0.620 & 0.740 & 0.350 \\
\tool{}-{lightGBM}     & \istrue & \istrue & 0.880 & \textbf{0.940} & 0.470 & 0.910 & \textbf{0.590} & 0.730 & 0.890 & 0.550 \\
\tool{}-{MLP}    & \istrue & \istrue & 0.900 & \textbf{0.940} & 0.550 & 0.940 & 0.550 & \textbf{0.750} & 0.890 & 0.600 \\
\tool{}-{XGBoost} & \istrue & \istrue & \textbf{0.920} & 0.920 & \textbf{0.790} & \textbf{0.980} & 0.369 & 0.660 & \textbf{0.920} & \textbf{0.650} \\
\bottomrule
\end{tabular}
}
\end{table*}

\section{Results}
\label{sec:results}
We evaluate our model on the two clinical prediction tasks: in-hospital mortality and 30-day readmission. 
Our experiments compare performance against strong baselines, including general and medical LLMs, and assess the value of key model components through ablation studies.

\subsection{Experimental Setup}
\label{sec:data}
\textbf{Datasets}. We use the MIMIC-III dataset~\cite{johnson2016mimic}, which includes structured and unstructured data for over 40,000 ICU patients. It includes structured data (demographics, vitals, labs, admissions, ICD-9-CM codes) and unstructured clinical text (physician notes, discharge summaries, radiology reports). For this study, we focus on physician-authored notes containing clinical reasoning, assessments, and treatment plans.
Here, we exclude discharge summaries and notes written after outcomes to prevent label leakage. Only $0.85\%$ of notes mention hospice, indicating rare explicit terminal indicators. However, a limitation of MIMIC-III for the readmission task is the inability to distinguish planned or elective readmissions and interfacility transfers, which may inflate the count of “avoidable” readmissions.

\noindent
\textbf{Biomedical Knowledge.}
We use abstracts from the annual PubMed Baseline dataset, comprising over 36 million biomedical citation records, to build a medical knowledge graph that enriches LLM input, and reduces hallucinations. We also incorporate UMLS \cite{bodenreider2004unified} to construct concept-centric subgraphs from EHR data.

\noindent
\textbf{Dataset Statistics.} We include a summary table in the Appendix, showing dataset statistics, indicatin moderate class imbalance for in-hospital mortality ($\sim 13\%$ positive) and severe imbalance for 30-day readmission ($\sim4\%$ positive). To prevent data leakage, train-test splits $(80:20)$ are patient disjoint, meaning that multiple visits from the same patient do not appear in both sets.

\noindent
\textbf{Baselines.} We compare against Claude 3.5 Sonnet~\cite{claude2024}, MedGemma~\cite{medgemma2024}, LLaMA3-Med~\cite{llama3med2024}, and KARE~\cite{jiangreasoning}, as well as structured-data models including logistic regression, tree-based models, and MLPs used in prior work on MIMIC III~\cite{harutyunyan2019multitask,lundberg2020local, purushotham2018benchmark}. All LLMs are evaluated in a zero-shot setting with the same patient-context prompt. KARE uses similar patient retrieval but lacks clinical notes. Our model incorporates retrieved notes for better context. Implementation details are in the Appendix.

\noindent
\textbf{Metrics.} We evaluate model performance using a comprehensive set of measures. AUROC and AUPRC capture discrimination ability across thresholds, while overall accuracy reflects the proportion of correct predictions. For class-specific evaluation, we report precision (positive predictive value, PPV), recall (sensitivity), negative predictive value (NPV), and specificity. These capture both the model’s ability to correctly identify positive cases (sensitivity, PPV) and its reliability on negatives (specificity, NPV). Finally, macro F1-score balances precision and recall across classes, ensuring that performance on the minority positive class is not overshadowed by the majority class. All metrics are computed on a held-out test set using standardized preprocessing to ensure comparability across models.

\subsection{Performance of \tool{}}

As we will later discuss in Section~\ref{sec:methods}, we explore different kinds of standard ML methods in $M_2$ for making the final prediction, using all the integrated inputs.
We refer to the corresponding algorithm as  \textbf{\tool{}-X}, where X represents Balanced-RF (balanced random forest), logistic regression, random forests, LSTM, LightGBM, MLP, or XGBoost.

\subsubsection{30-Day Readmission Prediction}

Readmission within 30 days is a highly imbalanced task, with only about $4\%$ positive cases in the dataset. This severe imbalance is reflected in the results, where most models achieve high accuracy and precision on the negative class but struggle with recall for the positive (readmitted) class.\\
As shown in Table~\ref{tab:merged_model_tasks}, our framework with Balanced Random Forest (\tool{}-BalancedRF) classifier achieves the highest AUROC $(0.845)$ and notably improves recall on positive cases to $0.79$, a crucial metric since identifying patients at risk of readmission is clinically imperative. The \tool{}-MLP model, while achieving the highest overall accuracy $(0.91)$ and macro F1 $(0.58)$, still attains a sensitivity of $0.28$ on positive cases, illustrating the persistent challenge in detecting rare events. 
Unstructured LLM-based baselines such as Claude-3.7-Sonnet, MedGemma, LLaMA3-Med, and KARE show substantially lower sensitivity for positives (below 0.3), suggesting that these models struggle to identify the minority class without further fine-tuning or domain-specific adaptation. 

In Table~\ref{tab:merged_model_tasks}, \tool{}-BalancedRF achieves a precision of 0.13, meaning that about one in eight patients flagged as high-risk were actually readmitted. Recall (sensitivity) is 0.79, indicating that the model correctly identifies nearly 8 out of 10 true readmissions—a clinically critical result. The F1-score of 0.55 reflects the balance between precision and recall. The Negative Predictive Value (NPV) is 0.99, showing that almost all patients predicted as low-risk were indeed not readmitted. Specificity is 0.80, meaning the model correctly classifies 8 in 10 patients who were not readmitted as low-risk.

\begin{figure}[h]
    \centering
\includegraphics[width=\linewidth]{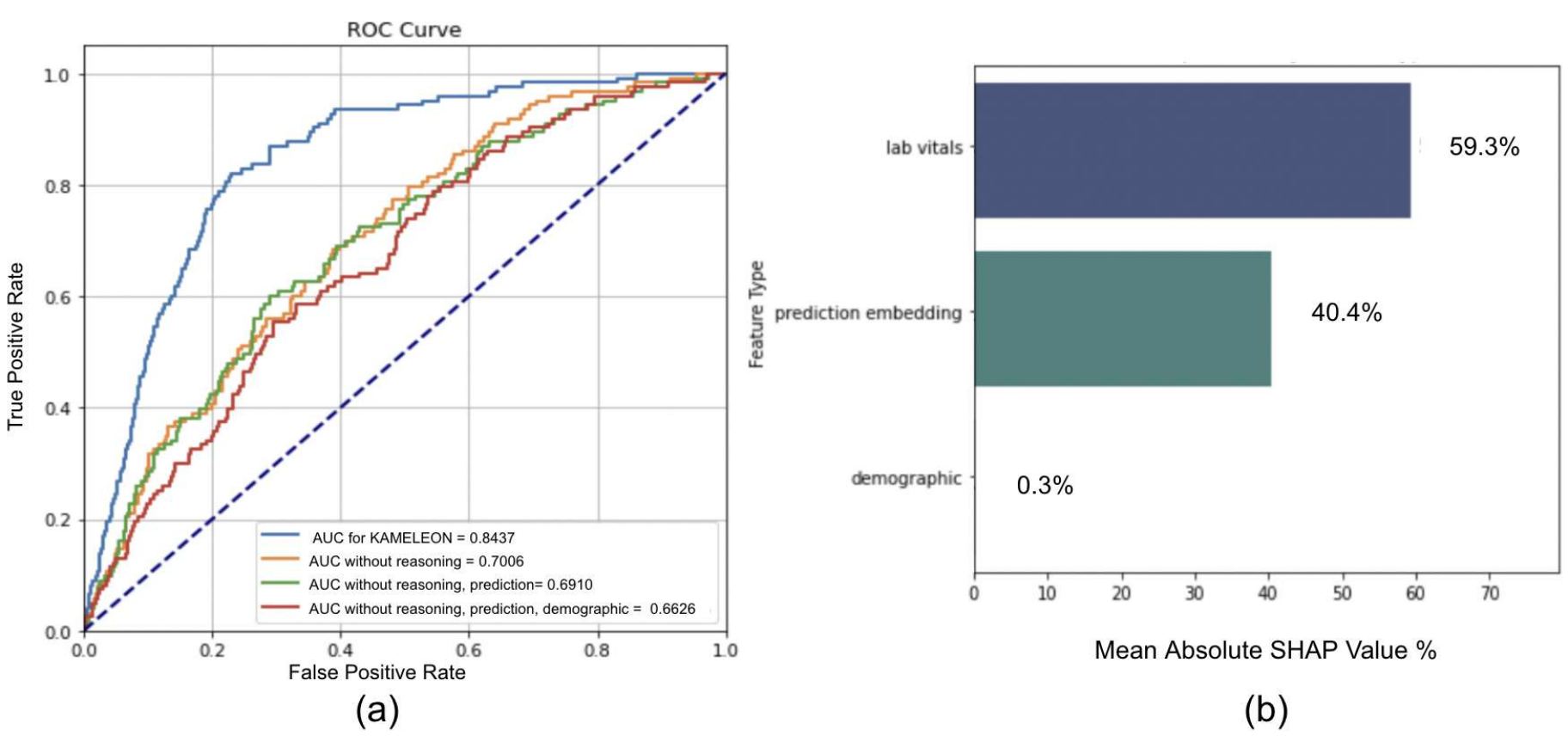}
    \vspace{-10pt}
    \caption{(a) \tool{} achieves the highest AUC for 30-day readmission when combining all features of \( M_1 \) and \( M_2 \), outperforming ablation variants. (b) SHAP analysis shows prediction embeddings from \( M_1 \) are key contributors.
}
\label{fig:ablation_readmission}
\end{figure}

To better understand this model’s behavior, we perform SHAP analysis to identify feature importance, and Figure~\ref{fig:ablation_readmission} indicates that the model relies primarily on prediction embeddings $(59.3\%)$ and lab/vital features $(40.4\%)$ for predicting 30-day readmission, highlighting the importance of multimodal inputs.

While prior studies like Morgan et al. (2019) \cite{morgan2019assessment} reported AUROCs up to 0.81 for readmission, and general models typically ranged from 0.61 to 0.73 \cite{dhalluin2020comparison, matheny2021development}, our multi-modal approach effectively captures complex clinical nuances.

\paragraph{Relative importance of different classes of inputs.}
We conduct an ablation study for the readmission task, where we retrain \tool{} after dropping different components--- the retraining and prediction models from $M_1$ and the demographics used in $M_2$
(Figure~\ref{fig:ablation_readmission} and Table~\ref{tab:ablation}). 
We find that the reasoning component output by $M_1$ is very significant, and affects multiple metrics beyond AUROC. 
In the full model, 
\tool{} achieves balanced performance (Acc = 0.80, Macro-F1 = 0.55, AUROC = 0.844, AUPRC = 0.147) with 
both high specificity (0.80) and sensitivity (0.77). 
When we drop the reasoning component,  sensitivity falls by over 80\% (falling to 0.13), 
and AUPRC is nearly halved, revealing strong bias toward the majority class; however, there is a gain in accuracy (rising to 0.92). 
Removing the reasoning component from \tool{} drops performance from 0.844 to 0.7, a 17\% decline in AUROC, highlighting the critical role of the finetuned LLM’s reasoning in risk prediction. Removing reasoning and prediction (output from $M_1$) causes intermediate degradation, with AUROC falling by $\sim18\%$, consistent with the loss of calibrated probability signals and semantic rationale. Eliminating reasoning together with demographics and prediction yields the sharpest overall decline, with Macro-F1 dropping by $\sim13\%$ and AUROC by more than $\sim22\%$ (to 0.663), confirming the complementary value of these components. These analyses collectively demonstrate that reasoning substantially improves minority-class detection: it boosts sensitivity and AUPRC by more than 
threefold compared to variants without it, while also preventing misleading accuracy gains driven solely by 
the dominant negative class. 
Figure~\ref{fig:ablation_readmission} visualizes these contributions, showing 
that reasoning, $M_1$’s prediction embedding, and demographics each add critical and non-redundant signals 
for accurate readmission risk estimation.

\vspace{-10pt}

\begin{table}[h]
\centering
\resizebox{\linewidth}{!}{%
\begin{tabular}{lccccccccc}
\hline
\textbf{Model} & \textbf{Acc} & \textbf{NPV} & \textbf{Precision (PPV)} & \textbf{Specificity} & \textbf{Sensitivity} & \textbf{Macro F1} & \textbf{AUROC} & \textbf{AUPRC}\\
\hline
\tool{} & 0.80 & 0.99 & 0.13 & 0.80 & 0.77 & 0.55 & 0.844 & 0.147\\


without  reasoning$_{M_1}$ & 0.94 & 0.96 & 0.07 & 0.97 & 0.06 & 0.52 & 0.699 & 0.078 \\

without demographic, reasoning$_{M_1}$, prediction$_{M_1}$ & 0.73 & 0.97 & 0.06 & 0.74 & 0.45 & 0.48 & 0.663 & 0.061 \\


without reasoning$_{M_1}$, prediction$_{M_1}$ & 0.74 & 0.97 & 0.09 & 0.75 & 0.55 & 0.50 & 0.691 & 0.062 \\

\hline
\end{tabular}%
}
\caption{Ablation study on Readmission}
\label{tab:ablation}
\end{table}

\paragraph{Sample Inference Output from $M_1$.}
Table~\ref{tab:inference_table} shows a sample output generated by the fine-tuned LLM for the 30-day hospital readmission prediction task. Instead of returning a raw binary value (0 or 1), the model is prompted to generate both a prediction and its reasoning based on the patient's diagnoses, procedures, medications, and comparisons with similar cases. This structured explanation allows the model to ground its prediction in a clinical context, improving reliability and interpretability over naive classification. By incorporating rationale into the output, the LLM demonstrates better alignment with real-world clinical decision-making.

\begin{table*}[h!]
\centering
\caption{Inference Results from Fine-Tuned LLM for 30-Day Readmission Prediction}
\begin{small}
\begin{tabular}{|p{17cm}|}
\hline
\textbf{Patient ID: 25070\_0} \newline
To predict whether Patient ID 25070\_0 will be readmitted within 30 days, we analyze the provided EHR context, focusing on conditions, procedures, and medications.

1. \textbf{Conditions}: The patient has multiple serious conditions, including non-Hodgkin's lymphoma and complications from a device, which are significant risk factors for readmission. However, they also have other conditions that may not be as acute (e.g., asthma, thyroid disorders).

2. \textbf{Procedures}: The patient has undergone several diagnostic procedures, including mastoidectomy, which could lead to complications but are not major surgeries. The patient did not have any major surgical interventions that typically increase readmission risk.

3. \textbf{Medications}: The patient is on a complex regimen including corticosteroids and other systemic medications. The presence of corticosteroids indicates a potential for complications, particularly in the management of their non-Hodgkin's lymphoma.

4. \textbf{Similar Patients}: The similar patients show that those with chronic obstructive pulmonary disease and heart failure have higher readmission rates. However, Patient ID 25070\_0 does not share the same chronic conditions with these patients.

5. \textbf{Overall Assessment}: Although the patient has several chronic and acute conditions, they do not have the same severity or complexity as the patients who have been readmitted. The absence of major recent surgeries and the complexity of their medication regimen suggests a lower risk of readmission.

\textbf{\# Prediction \#} \newline
0 \\
\hline
\textbf{Patient ID: 93900\_1} \newline
To predict whether Patient ID: 93900\_1 will be readmitted to the hospital within 30 days, we analyze the patient's EHR context, focusing on conditions, procedures, and medications.

1. \textbf{Conditions}: The patient has multiple chronic conditions, including congestive heart failure, diabetes mellitus with complications, and various gastrointestinal disorders. These conditions are known to increase the risk of exacerbations and complications, which are significant predictors of readmission.

2. \textbf{Procedures}: The patient has undergone significant procedures, such as debridement and upper gastrointestinal endoscopy. These procedures can lead to complications, necessitating close monitoring and follow-up care, which may contribute to readmission.

3. \textbf{Medications}: The patient is on a complex medication regimen, including beta-blockers and insulins. The complexity of this regimen can lead to adherence issues and potential side effects, which might complicate the patient's health status and increase the likelihood of readmission.

4. \textbf{Similar Patients}: The similar patients, such as Patient ID: 16994\_6, experienced complications from chronic diseases and were readmitted. This indicates a trend in patients with similar conditions and treatment regimens.

5. \textbf{Overall Risk Assessment}: The combination of chronic conditions, significant procedures, and complex medication regimens suggests a high risk of readmission due to potential exacerbations and complications.

Given this comprehensive analysis, it is reasonable to conclude that Patient ID: 93900\_1 will likely be readmitted to the hospital within 30 days based on the factors outlined above. \\
\hline
\end{tabular}
\end{small}
\label{tab:inference_table}
\end{table*}

\begin{figure}[h!]
    \centering
    \begin{subfigure}[b]{0.51\linewidth}
        \centering
      \includegraphics[width=\linewidth]{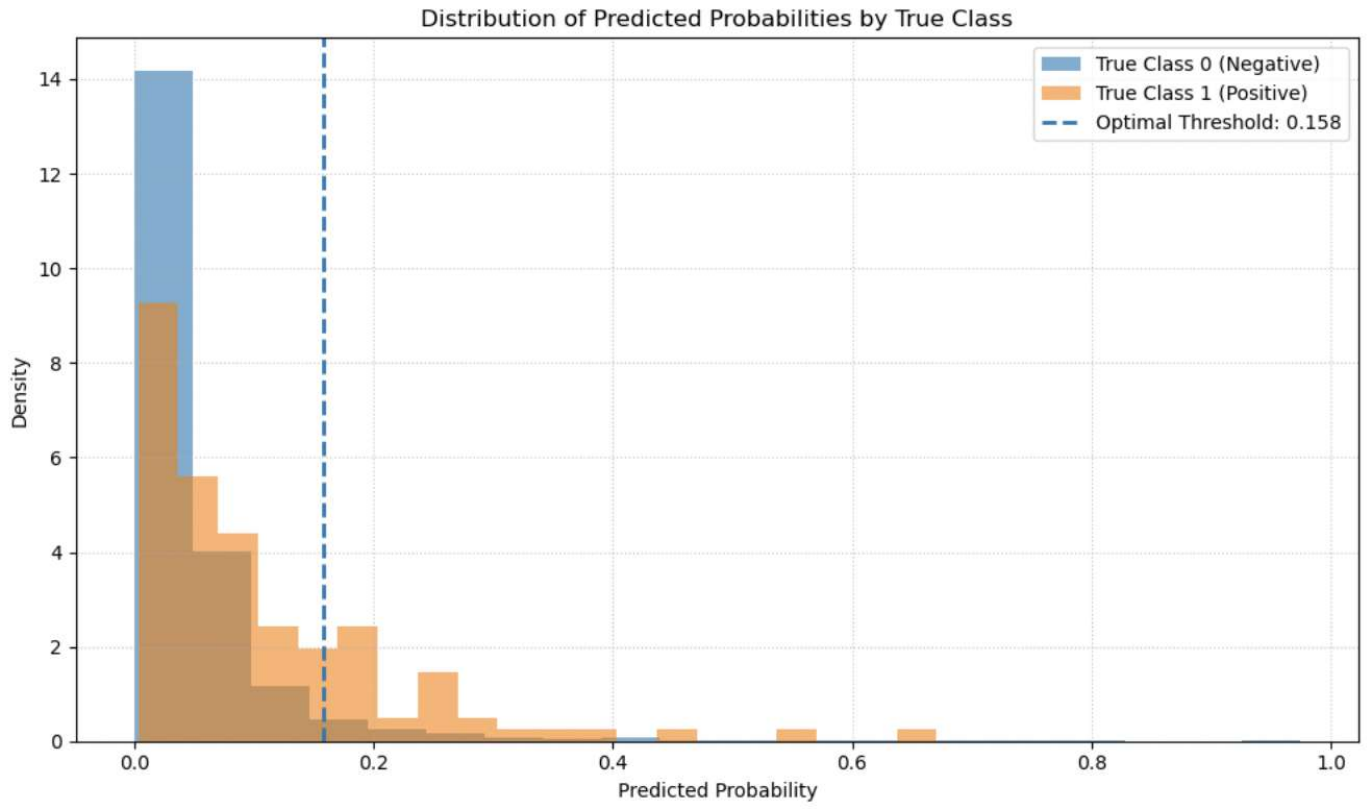}
        \caption{Distribution of predicted probabilities for readmission within 30 days, separated by true class labels. The vertical dashed line marks the optimal threshold (0.16) balancing sensitivity and specificity.}
        \label{fig:distribution_read}
    \end{subfigure}
    \hfill
    \begin{subfigure}[b]{0.45\linewidth}
        \centering
        \includegraphics[width=\linewidth]{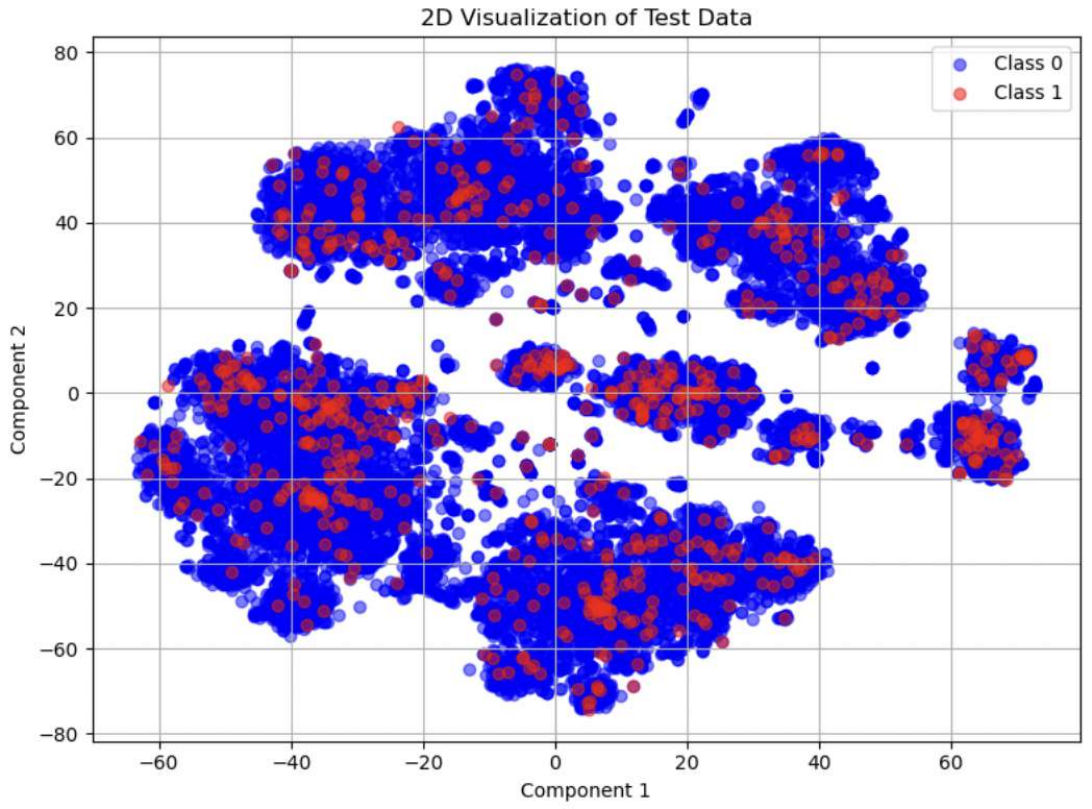}
        \caption{t-SNE embedding of the test dataset for the readmission in 30 days task, showing Class 0 (blue) and Class 1 (red).}
        \label{fig:tsne_read}
    \end{subfigure}
    \caption{Visualization of class separability and predicted probability distributions for the readmission in 30 days.}
    \label{fig:viz_read}
\end{figure}

\paragraph{Analysis of distribution (Readmission).} 
Figure~\ref{fig:distribution_read} illustrates the distribution of predicted probabilities for the readmission task. 
In the full model, the majority of negatives cluster near zero, while positives are shifted upward and 
concentrated above a low threshold of $\sim$0.16. This optimal threshold is chosen using Youden’s J statistic, which maximizes sensitivity and specificity. This separation, although not perfectly distinct, 
reflects the extreme class imbalance in the data: a very low cutoff is required to recover a reasonable 
fraction of positive cases. The overlap between classes explains the modest AUPRC values, since many 
positives still lie in regions dominated by negatives. Nevertheless, compared to ablated variants, the 
full model achieves tighter grouping of positives in the higher-probability region, resulting in better 
sensitivity and precision–recall trade-offs.

The t-SNE plot (Figure \ref{fig:tsne_read}) further visualizes patients' features on the held-out test set. In this space, readmitted 
patients do not form sharply separated clusters but instead appear partially embedded within the larger 
manifold of non-readmitted cases. The lack of clear separation reflects the difficulty of the task and 
the subtlety of signals driving readmission, where positives and negatives overlap substantially. Still, 
there is evidence of localized groupings of readmitted patients, suggesting that the model captures some 
latent patterns that distinguish higher-risk subgroups. This partial clustering is consistent with the 
modest AUPRC values: while the model cannot fully disentangle the classes, it is able to concentrate a 
portion of true positives in regions of elevated probability. Clinically, this underscores the challenge 
of predicting readmission but also highlights the value of identifying even partially coherent patient 
subgroups for targeted follow-up.

\newpage
\subsubsection{In-Hospital Mortality Prediction} 
\begin{figure}[h]
    \centering
    \begin{subfigure}[b]{0.44\linewidth}
        \centering
    \includegraphics[width=\linewidth]{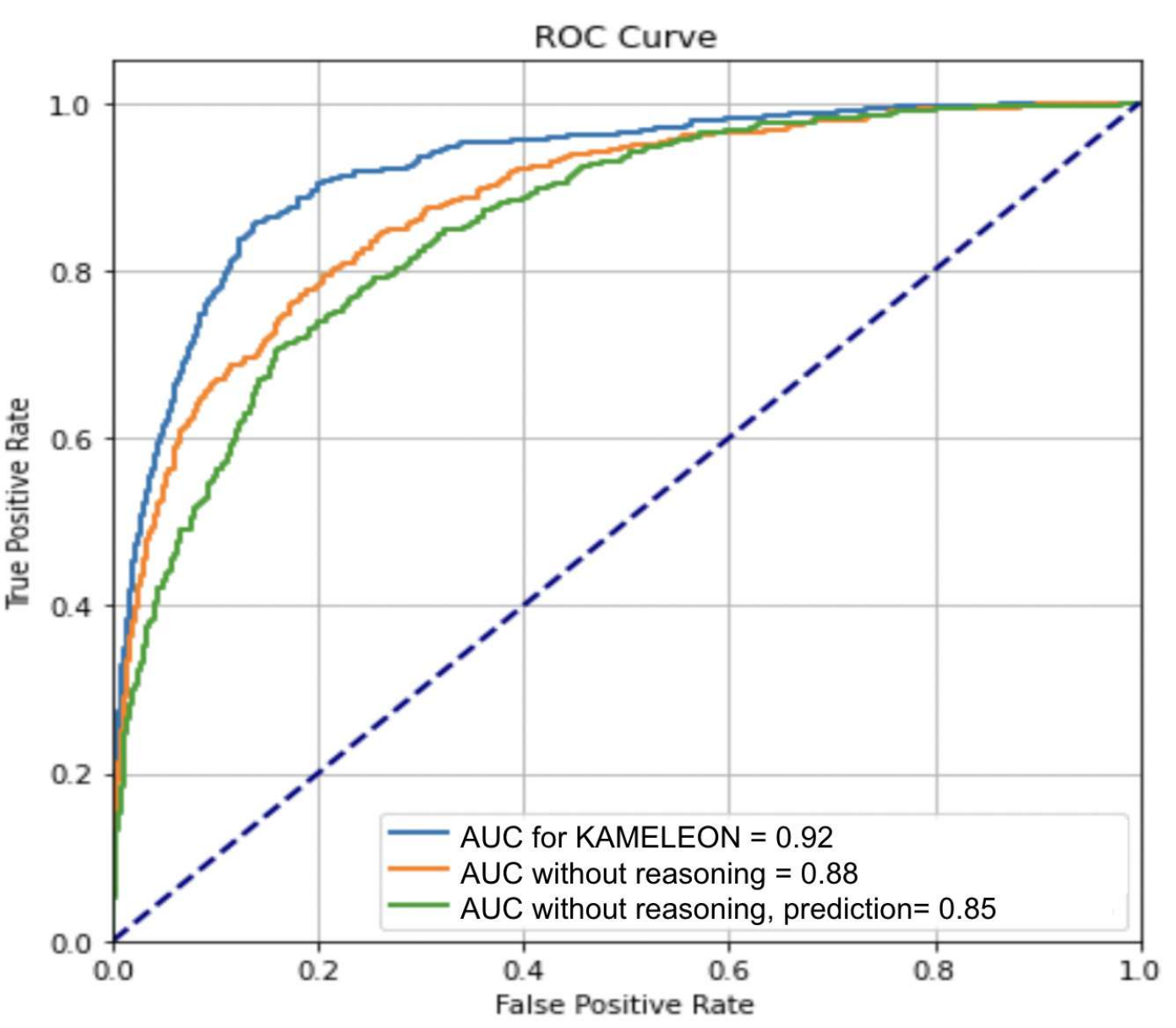}
        \caption{AUC for mortality prediction task}
        \label{fig:AUC_mort_}
    \end{subfigure}
    \hfill
    \begin{subfigure}[b]{0.54\linewidth}
        \centering
        \includegraphics[width=\linewidth]{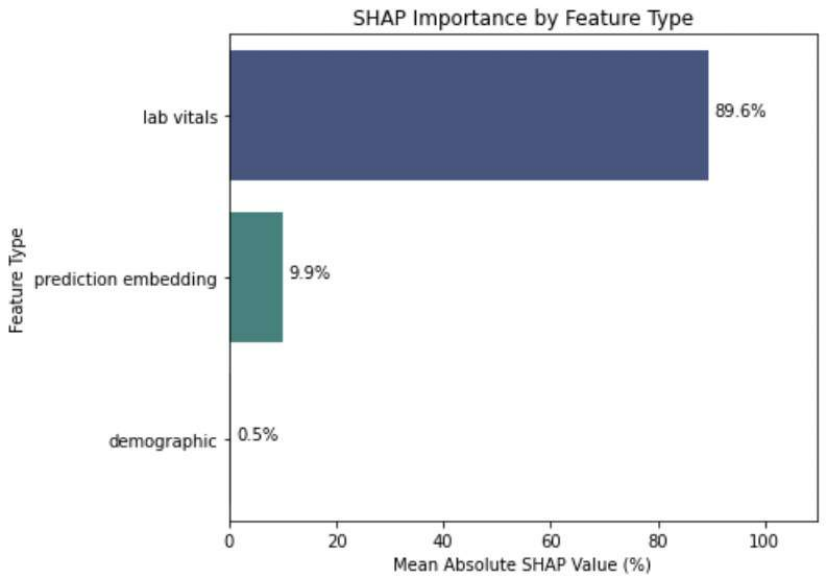}
        \caption{Importance of features for predicting mortality}
        \label{fig:shap_mort}
    \end{subfigure}
    \caption{Performance and feature importance for the mortality prediction task.}
    \label{fig:mortality_results}
\end{figure}

Mortality prediction is less imbalanced, with approximately $13\%$ positive cases. 
Table~\ref{tab:merged_model_tasks} reports that \tool{}-XGBoost and \tool{}-MLP models achieve high accuracy ($0.92$ and $0.90$ respectively) and AUROC ($0.92$ and $0.83$ respectively). \tool{}-XGBoost demonstrates strong performance for mortality prediction, achieving a high precision of $0.79$, meaning that most patients flagged as high-risk did not survive. It also attains a specificity of $0.98$ and an NPV of $0.92$, indicating that nearly all patients predicted as low-risk were indeed survivors. Furthermore, for a positive class that, while less imbalanced, is still a minority, the Area Under the Precision-Recall Curve (AUPRC) is a vital metric. Here, the \tool{} with $0.650$ sets the benchmark, significantly outperforming all other baselines. Unstructured models consistently yield the lowest performance for mortality prediction, AUROC values hover just above random chance (around $0.51 - 0.53$), and AUPRCs remain very low (max $0.125$), indicating a limited ability to discern between mortality and survival solely from clinical notes. SHAP results (Appendix Figure ~\ref{fig:shap_mort}) demonstrate that lab results and vital signs strongly drive predictions, with prediction embeddings playing a less critical role compared to readmission.
Overall, Table~\ref{tab:merged_model_tasks} shows that for both tasks, our multimodal model outperforms all individual structured and unstructured baselines across all metrics.

\paragraph{Relative importance of different classes of inputs.} 
The SHAP values (Figure~\ref{fig:shap_mort}) show that the prediction embedding has a smaller 
influence compared to labs and vitals, which contrasts with the readmission task. The AUROC curve 
(Figure~\ref{fig:AUC_mort_}) further illustrates that removing reasoning reduces AUROC from 
$\sim$0.92 in the full model to $\sim$0.88, a relative drop of about 4\%. This indicates that 
mortality prediction is comparatively easier, as a strong AUROC is retained even without reasoning. The lower class imbalance and the strong signal contained in lab values and diagnostic codes allow the model to capture patterns associated with terminal illness more directly. Nevertheless, reasoning still contributes by improving discrimination at the margin, capturing subtler risk factors that 
are less apparent in structured features alone.


\paragraph{Sample Inference Output from $M_1$.}
We show inference outputs for two patients on the mortality prediction task, generated by $M_1$, our finetuned LLM with reasoning in Table~\ref{tab:inference_table2_part2}, Appendix.

\begin{figure}[h!]
    \centering
    \begin{subfigure}[b]{0.5\linewidth}
        \centering
        \includegraphics[width=\linewidth]{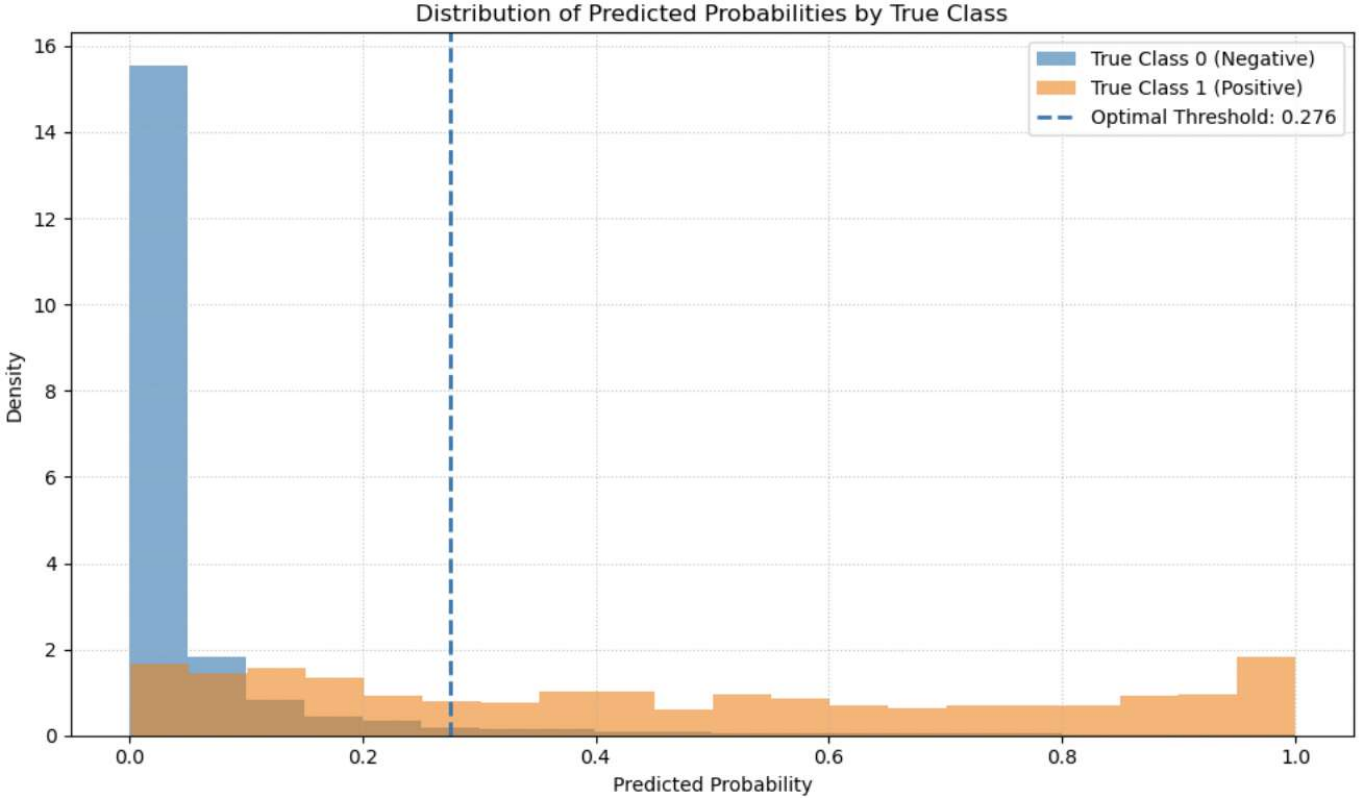}
        \caption{Distribution of predicted probabilities for in-hospital mortality within 30 days, separated by true class labels. The vertical dashed line marks the optimal threshold (0.276) balancing sensitivity and specificity.}
        \label{fig:distribution_mort}
    \end{subfigure}
    \hfill
    \begin{subfigure}[b]{0.47\linewidth}
        \centering
        \includegraphics[width=\linewidth]{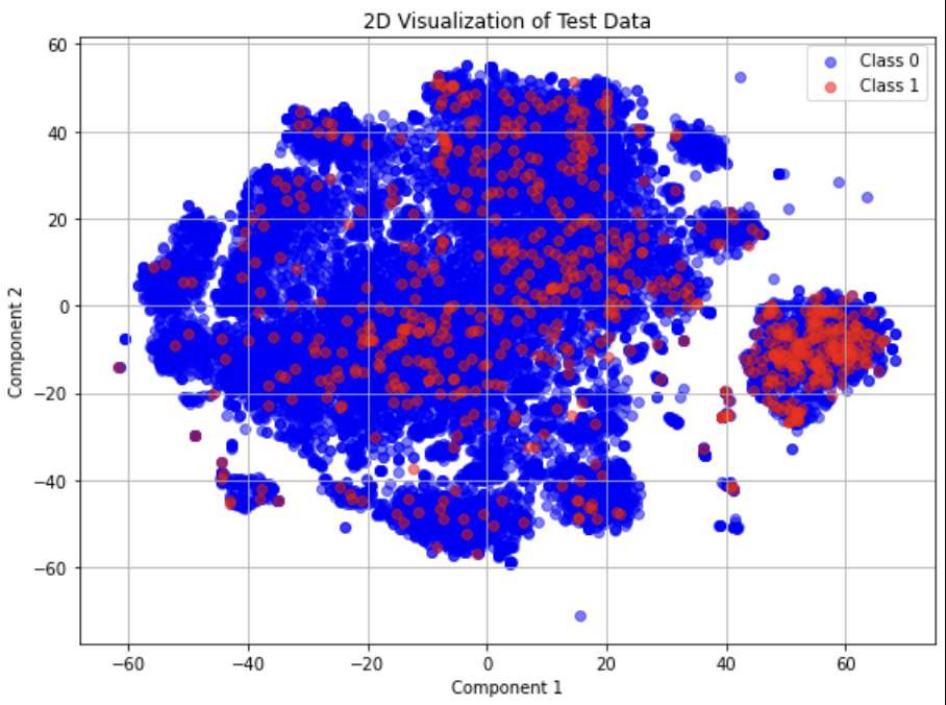}
        \caption{t-SNE embedding of the test dataset for the in-hospital mortality task, showing Class 0 (blue) and Class 1 (red).}
        \label{fig:tsne}
    \end{subfigure}
    \caption{Visualization of class separability and predicted probability distributions for the in-hospital mortality task.}
    \label{fig:viz_mortality}
\end{figure}

\paragraph{Analysis of distribution (In-Hospital Mortality).} Figure \ref{fig:distribution_mort} shows predicted probabilities for in-hospital mortality, separated by true class: survivors (blue) and deceased (red). The dashed line marks the optimal threshold (0.276) balancing sensitivity and specificity.

Most survivors cluster near zero probability, reflecting strong model confidence, while deceased cases spread across a wider range, showing prediction uncertainty. Overlap between classes causes some misclassifications, highlighting the challenge of predicting this rare event. Despite this, the clear separation and tight clustering of survivors demonstrate the model’s strong ability to distinguish between classes, supporting the usefulness of the selected threshold.


\noindent
\paragraph{Class Separability Analysis Using t-SNE (In-hospital Mortality Task).}

 Figure \ref{fig:tsne} shows a two-dimensional t-SNE embedding of the test data, where the samples are colored by their true class labels (Class 0 in blue and Class 1 in red). The visualization reveals that the majority of the data forms a consistent, structured manifold dominated by Class 0 points, with only a small fraction of Class 1 points distributed across the embedding. Notably, a compact cluster of Class 1 samples appears on the right-hand side, indicating localized patterns that can be exploited by advanced models. This structure suggests that although Class 1 is relatively sparse, it exhibits distinct feature signatures in specific regions, which the proposed \tool{}‑X model is designed to capture effectively, contributing to improved predictive performance.

\subsubsection{Incorporating additional patient context in $\textbf{M}_1$} 

Our unstructured model, $M_1$, extends KARE~\cite{jiangreasoning} by incorporating physician 
notes more explicitly into similar patients, enriching the knowledge graph; in contrast, KARE ~\cite{jiangreasoning} use only structured EHR data related to drugs, procedures, and conditions as context for similar patients, and the knowledge graph is constructed without considering patient conditions.
Our modification 
ensures that unstructured clinical narratives contribute to prediction alongside structured features, providing 
a stronger and more comparable baseline. 

Our strategy for adding physician notes leads to modest improvements in mortality and more substantial gains in readmission (Table~\ref{tab:merged_model_tasks}). For mortality prediction, AUROC and AUPRC improve by $\sim$2.3\% and $\sim$1.8\%, respectively. 
For readmission, the gains are stronger, with AUROC improving by $\sim$3.0\% and AUPRC by $\sim$2.1\%.
These results suggest that physician notes provide useful complementary signal, especially for readmission, where unstructured narratives capture behavioral, discharge-related risk factors less visible in structured EHR data.





\section{Discussion}
In this work, we introduce \tool{}, a novel framework that effectively integrates multimodal EHR data, including structured clinical features and unstructured physician notes, enhanced by knowledge-augmented LLM reasoning, for robust clinical risk prediction. 
Our two-stage architecture demonstrated superior performance on both 30-day readmission and in-hospital mortality prediction, with respect to multiple metrics, including the AUROC score.
\emph{\tool{} outperforms all prior baselines which only used one type of dataset (structured or unstructured), on most metrics for these two tasks, compared to prior work using the MIMIC-III dataset} (Table~\ref{tab:merged_model_tasks}).
Multiple types of standard ML methods have been used for these tasks, yet \tool{} demonstrates clear improvements across all evaluation metrics. None of the currently most powerful LLMs, including a medical LLM trained on clinical data, have comparable performance to \tool{}. The only exception is that the Llama3-Med42-8B model achieves a higher AUPRC for the readmission task; however, \tool{} significantly outperforms it across all other metrics.

We find that the reasoning component output by the LLM in $M_1$, which is used in $M_2$ by constructing an embedding, has high predictive power in both the tasks.
For the 30-day readmission task, the embedding constructed using the reasoning output by $M_1$ is very significant--- removing this component causes the AUROC to drop from $84.4\%$ to $68.7\%$.
This effect is much smaller in the case of the mortality prediction task, but not negligible,  dropping the AUROC from 0.92 to 0.88, when this component is dropped. 
This highlights the synergy achieved by combining these diverse modalities. 

This work underscores the significant potential of knowledge-augmented multimodal EHR modeling to enhance early intervention, optimize resource allocation, and improve patient care in complex clinical settings. 
While LLMs, including medical LLMs trained on specialized data, have a number of limitations in terms of accuracy and hallucinations, their reasoning outputs provide valuable predictive power.

Future work will focus on further validating \tool{}'s generalizability across diverse clinical settings and exploring its application to a wider range of predictive healthcare tasks.
Our framework can be easily extended to other clinical prediction tasks, especially those for which structured models have already been developed.
\tool{} can be applied for such tasks without any changes, and
we expect it will provide similar gains.

\noindent
\textbf{Scope and Promise for Social Impact.}
\tool{} offers a strong opportunity to reduce avoidable hospital readmissions, a major driver of morbidity, cost, and financial penalties \cite{fry2021frequent, joynt2013path, panagiotou2021association, zuckerman2017effect}. 
It provides real-time risk predictions for inpatients, enabling more effective discharge planning, case management, and post-acute care.
Our model uses real-time patient data to assess readmission risk prior to discharge, supporting individualized case management, discharge planning, census forecasting, and post-acute care coordination. 
By identifying high-risk patients, our model enables focused use of limited resources, improving efficiency and outcomes. 




\section{Methods}
\label{sec:methods}

\begin{table}[h]
\centering
\caption{Summary of notation used in the framework.}
\setlength{\tabcolsep}{4pt} 
\renewcommand{\arraystretch}{1.1} 
\begin{tabularx}{\linewidth}{| @{} l | X @{} |} 
\hline
\textbf{Symbol} & \textbf{Description} \\
\hline
$X^{\mathrm{struct}}$ & Structured clinical features \\
$X^{\mathrm{unstruct}}$ & Clinical free-text notes (e.g., physician notes) \\
$X^{\mathrm{demo}}$ & Demographic information \\
$X^{\mathrm{sim}}$ & Embeddings of similar patient notes \\
$\mathcal{G}, \mathcal{T}$ & Biomedical knowledge graph, triples \\
$H^{\mathrm{text}}$ & Unstructured text embedding \\
$H^{\mathrm{KG}}$ & Knowledge graph community summary embedding \\
$H^{\mathrm{LLM\_reasoning}}$ & LLM-generated reasoning with context \\
$D^{\mathrm{LLM}}_{\text{train/test}}$ & Augmented LLM training/test inputs \\
$f_{\mathrm{LLM}}$ & Fine-tuned large language model \\
$\hat{y}_{\text{reasoning}}, \hat{y}'$ & LLM-generated textual reasoning, output label\\
$f_{\mathrm{ML}}$ & Final machine learning classifier \\
$H^{\mathrm{concat}}$ & Concatenated features for $f_{\mathrm{ML}}$ \\
$\hat{y}_{\text{task}}$ & Final binary classification output \\
$\mathcal{L}_{\mathrm{LLM}}$ & LLM fine-tuning loss \\
$\mathcal{L}_{\text{task}}$ & Task-specific binary cross-entropy loss \\
\hline
\end{tabularx}
\vspace{-5pt}
\label{tab:notation_summary}
\end{table}

\subsection{Background}
\noindent
\textbf{Clinical Datasets.}
Clinical data used in this study includes both structured and unstructured data collected across patient visits, capturing the patient's condition over time.

\begin{itemize}
    \item Structured data includes standardized fields such as lab results, vital signs, demographic attributes (e.g., age, sex, ethnicity). These are typically numeric or categorical and readily usable for statistical modeling.

    \item Unstructured data consists of free-text clinical documentation such as physician notes, as well as patient conditions, diagnoses, and prescribed medications. 
\end{itemize}

\noindent
\textbf{Additional datasets.} 
PubMed and UMLS (Section~\ref{sec:data}).


\noindent
\textbf{Problem statements.}
To demonstrate the effectiveness of our method, we study two popular clinical tasks: in-hospital mortality prediction and 30-day readmission prediction, e.g.,~\cite{jiangreasoning, cai2016real}. 
We define these problems formally after introducing some notation.

\noindent
\textbf{Notation:} We use $v_i$ to denote a hospital visit by a patient. For each visit $v_i$, the patient is associated with a set of medical data, $\mathcal{D}_i = \mathcal{D}^{\mathrm{struct}}_i \cup \mathcal{D}^{\mathrm{unstruct}}_i$, comprising both \textit{structured information} $\mathcal{D}^{\mathrm{struct}}_i$ (e.g., codes, vitals, lab results) and \textit{unstructured information} $\mathcal{D}^{\mathrm{unstruct}}_i$ (e.g., clinical free-text notes). Our goal is to build a model $f_{\theta}$ that predicts a patient's target status based on their historical visit information, specifically, $\hat{y}_i = f_{\theta}(\mathcal{D}_i)$.

\noindent
\textbf{30-day Readmission Problem.} The objective is to determine whether the patient is readmitted to the hospital within 30 days following discharge from visit $v_i$. We define the readmission indicator $y_{v_i}^{readm}$ as:
{\small
\[
y_{v_i}^{readm} =
\begin{cases}
1, & \text{if patient is readmitted within 30 days of visit } v_i \\
0, & \text{otherwise}
\end{cases}
\]
}
The goal is to develop a predictive model that estimates $y_{v_i}^{readm}$ using all structured and unstructured data from $v_i$.

\noindent
\textbf{Mortality Prediction Problem.} Given the complete set of information for a visit $v_i$, the objective is to determine the patient's in-hospital mortality status, denoted as $y_{v_i}^{mort}$, where:
\[
y_{v_i}^{mort} =
\begin{cases}
0, & \text{if the patient survived the visit} \\
1, & \text{if the patient died during the visit}
\end{cases}
\]
The goal is to develop a predictive model that accurately estimates $y_{v_i}^{mort}$ based on all available structured and unstructured clinical data associated with visit $v_i$.

\subsection{\tool{} Framework}\label{sec:framework}
 We propose a hybrid framework, \tool{}, that integrates multimodal EHR data, including structured clinical components and unstructured physician notes (we will provide more details in Section~\ref{sec:data}), and external biomedical knowledge to predict two key clinical outcomes: in-hospital mortality and 30-day readmission. 
As shown in Figure~\ref{fig:framework}, \tool{} consists of two components: 
(1) an unstructured encoder \( M_1 \) (Section ~\ref{sec:unstruct}) that processes clinical notes and retrieves relevant biomedical knowledge using a PubMed-derived graph and knowledge-augmented reasoning, and outputs a prediction, along with its reasoning; and 
(2) a structured encoder \( M_2 \) (Section ~\ref{sec:struct}) that combines multiple time-series corresponding to vitals and tabular datasets (labs, medications, etc), along with the outputs from \( M_1 \) (i.e., both the prediction and the embedding associated with the reasoning it produces) with static features for downstream prediction.
The notations used in Section \ref{sec:framework}, Algorithm \ref{alg:alg1}, and \ref{alg:alg2} are explained in Table \ref{tab:notation_summary}.
\smallskip



        

\subsubsection{Unstructured Model, $\mathbf{M}_1$}
\label{sec:unstruct}
For each hospital visit \( v_i \), we collect physician-authored clinical notes and extract entities like conditions, procedures, and medications. To enrich context, we use PubMed literature parsed into knowledge triples (subject–relation–object) via an LLM-based extraction pipeline. We retain only triples that appear across patient visits. These triples form a biomedical knowledge graph, serving as an auxiliary source to support LLM reasoning and diagnosis.


\noindent
\textbf{Generating Context.}
The first step of the framework is generating patient context. We use EHR data, including physicians’ notes, patient conditions, prescribed medications, and procedures. Since physician notes can be lengthy and exceed the context window of a small, locally fine-tuned LLM, we summarize them using an LLM. This approach addresses \textit{Challenge 2 (Long Context)}.

\noindent
\textbf{Knowledge Graph Retrieval.}
To introduce domain-level medical knowledge, we build a biomedical knowledge graph (KG) by combining the Unified Medical Language System (UMLS)\cite{bodenreider2004unified}, PubMed abstracts, and large language model (LLM)-generated entity-relation triples. UMLS provides standardized biomedical concepts and relationships, and the entity-relation triples are structured facts extracted by the LLM in the form \((\text{entity}_1, \text{relation}, \text{entity}_2)\), capturing semantic connections.
We apply the \textit{Leiden algorithm} for community detection \cite{leiden}, which partitions the KG into semantically coherent and well-connected subgraphs. After clustering, we employ a separate LLM to generate a textual summary for each cluster. These summaries are produced by reasoning over the relationships among entities within each cluster, capturing the latent biomedical semantics encoded in the graph. Each summary serves as a high-level abstraction of the biomedical concepts and interactions within a subgraph. We embed these summaries using SentenceTransformer (MiniLM-L6-v2) \cite{reimers_sentence} and retrieve the most relevant ones for each patient by computing semantic similarity with the embedded patient context.
This process directly addresses \textit{Challenge 3} \textit{(Specialized Medical Domains)} by enriching patient context with structured, domain-specific knowledge, improving the model’s understanding of specialized medical terminology.

\noindent
\textbf{Finding Similar Patients.}
We provide additional context, by retrieving semantically similar patient visits using precomputed visit-level embeddings and a similarity index using the FAISS \cite{douze2024faiss} library with inner-product search on L2-normalized embeddings, which effectively approximates cosine similarity. 
For each target patient visit, we retrieve the top 50 most similar patients while excluding self-matches and other visits from the same individual. Each retrieved patient is scored by similarity, and we filter them into positive and negative cohorts based on matching or non-matching ground truth labels (e.g., readmission vs. no readmission). The final output includes the top-k positive and negative similar patients (with k=1,2).
Unlike KARE \cite{jiangreasoning}, we also provide the physician notes of the retrieved similar patients, enabling the language model to leverage more clinical context when assessing patient risk.

\noindent
\textbf{Reasoning Module.}
In this module, we prepare inputs to fine-tune the LLM for clinical prediction. For each patient visit, we create a prompt with the patient’s context with the top-k most similar cases retrieved earlier. These similar cases guide the model by highlighting patterns in clinically comparable scenarios. We also add biomedical knowledge summaries from clustered subgraphs of a PubMed knowledge graph, providing literature-based context. Combining patient data, historical cases, and domain knowledge, we fine-tune the LLM to produce task-specific predictions with interpretable reasoning, supporting each outcome.

\noindent
\textbf{Finetuning LLM.} We fine-tune a \textit{LLaMA-3 8B} model using the Unsloth framework \cite{meta2024llama3, unsloth}, which enables memory-efficient training via 4-bit quantization and Low-Rank Adaptation (LoRA) \cite{hu2021lora}. Prediction tasks are framed as instruction-following using Alpaca-style prompts with task description, patient context, and optional justification. Each prompt combines clinical notes, retrieved similar cases, and biomedical knowledge summaries. The model is trained via supervised learning to generate both predictions and reasoning. We use limited training steps with gradient accumulation and sequence lengths up to $8192$ tokens. Unlike KARE, which trains larger models with higher compute, our method uses smaller, quantized models to reduce computational cost while maintaining interpretability and performance. 
Algorithm~\ref{alg:alg1} outlines the training and inference procedures of $M_1$.
Additionally, an example prompt and its overall structure are provided in the Appendix Section \ref{M_1:appendix}.

\subsubsection{Structured Data Encoder, $\mathbf{M}_2$}
\label{sec:struct}
We extract structured features from the patient's visit history during each stay, including:
\begin{itemize}
    \item \textbf{Time-varying variables.} We extract hourly vitals and diagnoses in MIMIC-III, including heart rate, systolic and diastolic blood pressure, mean blood pressure, oxygen saturation, Glasgow Coma Scale (GCS) scores, glucose level, respiratory rate, temperature, weight, and pH.
    
    \item \textbf{Static metadata.} Demographic and admission features include gender, ethnicity, admission type, location, insurance, language, religion, and hospital outcome.

    \item \textbf{Diagnoses, procedures, medications.} ICD-9 codes, drug names are encoded via one-hot or counts. We compute binary indicators for key comorbidities (e.g., sepsis, infection, organ failure, dementia, cancer, diabetes).
\end{itemize}






\noindent
\textbf{Structured Data Preprocessing.}
The structured data is first transformed using a discretization step to enforce uniform temporal resolution and impute missing values. This is followed by normalization using precomputed mean and standard deviation statistics over the continuous variables.
\begin{algorithm}[h]
\caption{Unstructured Data Encoder, $M_1$}
\label{alg:alg1}
\small
\begin{algorithmic}[1]
\REQUIRE{Training data: \( \mathcal{D}_{\text{train}}^{\text{unstruct}} = \{ (X^{\mathrm{unstruct}}_{\text{train},i}, y_{\text{task},\text{train},i}) \}_{i=1}^{N_{\text{train}}} \)}
\REQUIRE Test data: \( \mathcal{D}_{\text{test}}^{\text{unstruct}} = \{ X^{\mathrm{unstruct}}_{\text{test},j} \}_{j=1}^{N_{\text{test}}} \)
\ENSURE Intermediate prediction \( \hat{y}' \in [0, 1]^{N_{\text{test}}} \), Reasoning \( \hat{y}_{\text{reasoning}} \)

\vspace{0.5em}
\noindent\textbf{Phase 1: Pre-processing \& LLM Input Preparation}
\FOR{each sample \( x^{\mathrm{unstruct}} \) in \( \mathcal{D}_{\text{train}}^{\text{unstruct}} \cup \mathcal{D}_{\text{test}}^{\text{unstruct}} \)}
    \STATE \( H^{\mathrm{text}} \gets \texttt{Enc}_{\mathrm{text}}(x^{\mathrm{unstruct}}) \)
    \STATE \( X^{\mathrm{sim}} \gets \texttt{RetrieveSimilarPatients}(x^{\mathrm{unstruct}}, \mathcal{G}) \) \COMMENT{via FAISS}
    \STATE \( H^{\mathrm{sim}} \gets \texttt{Aggregate}(\{ \texttt{Enc}_{\mathrm{text}}(X^{\mathrm{sim}}_j) \}_{j=1}^M) \)
    \STATE \( \mathcal{T} \gets \texttt{ExtractTriples}(x^{\mathrm{unstruct}}) \cap \texttt{ExtractTriples}(\texttt{PubMed}) \)
\ENDFOR
\STATE \( \mathcal{G} \gets \texttt{BuildKnowledgeGraph}(\mathcal{T}) \)
\STATE \( \{ \mathcal{C}_k \} \gets \texttt{ClusterGraph}(\mathcal{G}) \)
\STATE \( H^{\mathrm{KG}} \gets \texttt{CommunitySummary}_{\mathrm{KG}}(\{ \mathcal{C}_k \}, \texttt{LLM}_1) \) 
\vspace{0.5em}
\FOR{each sample \( x^{\mathrm{unstruct}} \) in \( \mathcal{D}_{\text{train}}^{\text{unstruct}} \)}
    \STATE \( H^{\text{LLM\_reasoning}} \gets \texttt{GenerateReasoning}(H^{\mathrm{text}}, H^{\mathrm{KG}},\)  \(H^{\mathrm{sim}},y_{\text{true}}) \)
\ENDFOR

\STATE Construct \( D_{\text{train\_X}}^{\text{LLM}} \gets \{ H^{\mathrm{text}}, H^{\mathrm{KG}}, H^{\mathrm{sim}}, H^{\text{LLM\_reasoning}} \} \)
\STATE Construct \( D_{\text{test\_X}}^{\text{LLM}} \gets \{ H^{\mathrm{text}}, H^{\mathrm{KG}}, H^{\mathrm{sim}} \} \)

\vspace{0.5em}
\noindent\textbf{Phase 2: Fine-tuning LLM Model (\( f_{\mathrm{LLM}} \))}
\STATE Initialize optimizer for \( f_{\mathrm{LLM}} \)
\FOR{epoch = 1 to \textit{NumEpochsLLM}}
    \FOR{each batch \( \{ h^{\mathrm{text}}, h^{\mathrm{KG}}, h^{\mathrm{sim}}, h^{\text{LLM\_reasoning}} \} \) in \( D_{\text{train\_X}}^{\text{LLM}} \)}
        \STATE \( (\hat{y}'_{\text{batch}}, \hat{y}_{\text{reasoning},\text{batch}}) \gets f_{\mathrm{LLM}}(h^{\mathrm{text}}, h^{\mathrm{KG}},h^{\mathrm{sim}}, \)
        \( h^{\text{LLM\_reasoning}}) \)
        \STATE \( \mathcal{L}_{\mathrm{LLM}} \gets \texttt{CrossEntropy}(\hat{y}'_{\text{batch}}, y_{\text{true},\text{batch}}) + \mathcal{L}_{\text{auxiliary}}(h^{\mathrm{text}}, h^{\mathrm{sim}}) \)
        \STATE Backpropagate \( \mathcal{L}_{\mathrm{LLM}} \); update \( f_{\mathrm{LLM}} \)
    \ENDFOR
\ENDFOR

\vspace{0.5em}
\noindent\textbf{Phase 3: Inference with Fine-tuned LLM}
\FOR{each test sample in \( D_{\text{test\_X}}^{\text{LLM}} \)}
    \STATE \( (\hat{y}', \hat{y}_{\text{reasoning}}) \gets f_{\mathrm{LLM}}(H^{\mathrm{text}}, H^{\mathrm{KG}}, H^{\mathrm{sim}}) \)
\ENDFOR

\STATE \textit{return} \( (\hat{y}', \hat{y}_{\text{reasoning}}) \)
\end{algorithmic}
\end{algorithm}

\noindent
\textbf{Incorporating LLM Output.}
To augment the structured input, we include the LLM’s prediction and its tokenized reasoning. A Word2Vec \cite{mikolov2013efficient} model is trained on this reasoning to generate embeddings. Each paragraph is represented by averaging its word vectors, producing a fixed-size vector. This reasoning vector is concatenated with the output label to form a feature vector for each visit.

\noindent
\textbf{Final Integration.}
The LLM-derived vector is merged with structured input features to create a unified representation, directly addressing \emph{Challenge 1 (Multi-modal Information)}. To reduce dimensionality and suppress noise from high-dimensional embeddings, we apply Principal Component Analysis (PCA) to the combined feature vector. 
\smallskip

\subsubsection{Training}
\label{sec:train}
We follow a two-stage training procedure. First, we fine-tune the unstructured text encoder \( M_1 \) using instruction-style prompts built from clinical notes, retrieved similar cases, and external biomedical knowledge. After fine-tuning, we perform \textit{Final Integration} and use the outputs of \( M_1 \) as input features to train \( M_2 \) for final prediction. 

In our experiments, we benchmark several ML models for \( M_2 \) as \textbf{\tool{}-X}, where X represents logistic regression, balanced random forests (BRF), LSTM, LightGBM, MLP, or XGBoost, selected for its effectiveness in capturing clinical patterns. 
For MLP, we use weighted binary cross-entropy loss\\

\begin{equation*}
\mathcal{L}_{\text{WBCE}} = -\frac{1}{N} \sum_{i=1}^{N} \left[ w_1 \cdot y_i \cdot \log(\hat{y}_i) + w_0 \cdot (1 - y_i) \cdot \log(1 - \hat{y}_i) \right]
\label{eq:wbce}
\end{equation*}

\noindent
where \(w_1\) and \(w_0\) are the positive and negative class weights, respectively, used to address class imbalance by giving more emphasis to the minority class. We further use Synthetic Minority Over-sampling Technique (SMOTE) \cite{chawla2002smote} to mitigate class imbalance, addressing \emph{Challenge 4 (Highly imbalanced data)}. 

\noindent
For \tool{}-BRF model, which is designed to handle class imbalance. Here, each decision tree is trained on a bootstrapped sample drawn by under-sampling the majority class and combining it with all minority class examples, ensuring balanced class proportions at the tree level. Each split is chosen to minimize the \textit{Gini impurity}:

\[
\text{Gini} = 1 - \sum_{c \in \{0,1\}} p_c^2,
\] where $p_c$ denotes teh proportion of class $c$ in a given node. The balanced sampling overcomes the issue with the dominance of the majority class and improves sensitivity to rare outcomes. Final predictions are obtained by aggregating probabilities across trees. 

The complete training and inference pipeline for $M_2$ is outlined in Algorithm~\ref{alg:alg2}.

\begin{algorithm}[h]
\caption{Structured Data Encoder, $M_2$}
\label{alg:alg2}
\small
\begin{algorithmic}[1]
\REQUIRE Structured training data: \( \mathcal{D}_{\text{train}}^{\text{struct}} = \{ (X^{\mathrm{struct}}_{\text{train},i}, X^{\mathrm{demo}}_{\text{train},i}, y_{\text{task},\text{train},i}) \}_{i=1}^{N_{\text{train}}} \)
\REQUIRE Structured test data: \( \mathcal{D}_{\text{test}}^{\text{struct}} = \{ (X^{\mathrm{struct}}_{\text{test},j}, X^{\mathrm{demo}}_{\text{test},j}) \}_{j=1}^{N_{\text{test}}} \)
\REQUIRE From Algorithm~\ref{alg:alg1}: \( (\hat{y}'_{\text{train}}, \hat{y}_{\text{reasoning, train}}), (\hat{y}'_{\text{test}}, \hat{y}_{\text{reasoning, test}}) \)
\ENSURE Final prediction \( \hat{y}_{\text{task}} \in [0,1]^{N_{\text{test}}} \)

\vspace{0.5em}
\noindent\textbf{Phase 1: Training Final ML Model (\(f_{\mathrm{ML}}\))}
\STATE Initialize optimizer for \( f_{\mathrm{ML}} \)
\STATE Train Word2Vec model on all \( \hat{y}_{\text{reasoning, train}} \)

\STATE Initialize empty dataset \( \mathcal{D}_{\text{ML\_train\_final}} \)
\FOR{each sample \( (x^{\mathrm{struct}}, x^{\mathrm{demo}}, y_{\text{true}}, \hat{y}', \hat{y}_{\text{reasoning}}) \) in training set}
    \STATE \( emb^{\text{reasoning}} \gets \texttt{Word2Vec}(\hat{y}_{\text{reasoning}}) \)
    \STATE \( H^{\mathrm{concat}} \gets \texttt{Concat}(x^{\mathrm{struct}}, x^{\mathrm{demo}}, \hat{y}', emb^{\text{reasoning}}) \)
    \STATE Add \( (H^{\mathrm{concat}}, y_{\text{true}}) \) to \( \mathcal{D}_{\text{ML\_train\_final}} \)
\ENDFOR

\FOR{epoch = 1 to \textit{NumEpochsML}}
    \FOR{each batch \( \{ H^{\mathrm{concat}}, y_{\text{true}} \} \) in \( \mathcal{D}_{\text{ML\_train\_final}} \)}
        \STATE \( \hat{y}_{\text{task}} \gets f_{\mathrm{ML}}(H^{\mathrm{concat}}) \)
        \STATE \( \mathcal{L}_{\text{task}} \gets \texttt{WeightedBCE}(\hat{y}_{\text{task}}, y_{\text{true}}) \)
        \STATE Backpropagate \( \mathcal{L}_{\text{task}} \); update model parameters
    \ENDFOR
\ENDFOR

\vspace{0.5em}
\noindent\textbf{Phase 2: Inference on Test Set}
\STATE Initialize empty list \( \hat{y}_{\text{task}} \)
\FOR{each test sample \( (x^{\mathrm{struct}}, x^{\mathrm{demo}}, \hat{y}', \hat{y}_{\text{reasoning}}) \)}
    \STATE \( emb^{\text{reasoning}} \gets \texttt{Word2Vec}(\hat{y}_{\text{reasoning}}) \)
    \STATE \( H^{\mathrm{concat}} \gets \texttt{Concat}(x^{\mathrm{struct}}, x^{\mathrm{demo}}, \hat{y}', emb^{\text{reasoning}}) \)
    \STATE \( \hat{y} \gets f_{\mathrm{ML}}(H^{\mathrm{concat}}) \)
    \STATE Append \( \hat{y} \) to \( \hat{y}_{\text{task}} \)
\ENDFOR

\STATE \textit{return} \( \hat{y}_{\text{task}} \)
\end{algorithmic}
\end{algorithm}
\vspace{-5pt}

\clearpage
\bibliographystyle{naturemag}
\bibliography{nc}
\appendix
\clearpage

\appendix

\section{Additional Details on Dataset}
We use the Medical Information Mart for Intensive Care III (MIMIC-III) dataset \cite{johnson2016mimic}, a large-scale, publicly available critical care database comprising deidentified health-related data associated with over $40,000$ patients who were admitted to the intensive care units (ICUs) of Beth Israel Deaconess Medical Center between $2001$ and $2012$. The dataset was curated to support clinical and epidemiological research and has since become a cornerstone resource in machine learning for healthcare.

MIMIC-III offers a comprehensive view of patient health trajectories by incorporating a diverse range of structured and unstructured data sources. Structured data include demographic information (e.g., age, gender, ethnicity), vital signs recorded at regular intervals (e.g., heart rate, blood pressure, respiratory rate, oxygen saturation), laboratory test results (e.g., glucose, pH, hematocrit), and hospital admission details (e.g., insurance type, admission location, length of stay, discharge disposition). Diagnosis and procedure codes are also available via ICD-9-CM, allowing for condition-based patient cohort extraction. 

In addition to structured clinical variables, MIMIC-III contains extensive \textit{unstructured clinical text}, such as physician notes, radiology reports, discharge summaries, Rehab Services, Echo, ECG , Case Management, General, Social Work, Pharmacy, Consult, and Nursing progress notes. These textual artifacts provide rich narrative context that captures clinical reasoning, impressions, and temporal evolution of patient conditions—information often not fully represented in structured fields.

For this study, we primarily focus on the unstructured physician notes that are authored during and after patient examinations. These notes include clinical assessments, differential diagnoses, treatment plans, and observations made by attending or resident physicians. We extract all available physician notes for each patient and temporally align them using the associated charted timestamps. To ensure semantic fidelity and medical relevance, a subset of the extracted notes is manually reviewed and annotated in collaboration with licensed physicians, enabling accurate downstream processing and interpretation by our model.

\subsection{Statistics of Dataset}

The datasets for 30-day readmission and in-hospital mortality prediction are highly imbalanced, as only a small fraction of patients who are discharged are readmitted within 30 days. Table \ref{tab:dataset_stats} shows the percentage of positive cases for both tasks in the training and test datasets.
\begin{table}[h]
\centering
\caption{Dataset statistics for mortality and readmission prediction tasks. Positive = target outcome occurred.}
\label{tab:dataset_stats}
\resizebox{0.5\linewidth}{!}{%
\begin{tabular}{|l|c|c|c|}
\toprule
\textbf{Task} & \textbf{Split} & \textbf{\#Samples} & \textbf{\% Positive} \\
\midrule
In-Hospital Mortality & Train & $17903$
 & $13.53\%$ \\
& Test  & $3236$ & $11.55\%$ \\
\midrule
Readmission in 30 days & Train & $10031$ & $4.01\%$ \\
& Test  & $2425$ & $3.80\%$ \\
\bottomrule
\end{tabular}
}
\end{table}

\subsection{Features}
Table~\ref{tab:mimic3_features} outlines the structured and unstructured features used to train our framework. We incorporate all available patient, and visit-level information, including vitals, lab results, diagnosis and procedure codes, medications, and clinical notes, to provide a comprehensive representation of each patient's condition. This multimodal input enables the model to learn both high-level clinical states and subtle variations across patients.

Table~\ref{tab:icd_categories} details the ICD-9 code groupings used to derive comorbidity indicators. These comorbidities are included as structured features, allowing the model to capture critical clinical context such as the presence of chronic diseases (e.g., diabetes, cardiovascular disease) or acute conditions (e.g., sepsis, organ dysfunction). We also found that incorporating diagnosis type and the frequency of related conditions (e.g., number of cancer-related diagnoses) further improved prediction performance, suggesting that leveraging structured clinical coding hierarchies contributes meaningful signal to patient outcome modeling.

\begin{table*}[ht]
\centering
\caption{Structured vs. Unstructured Features in the MIMIC-III Dataset}
\label{tab:mimic3_features}
\begin{tabular}{|l|l|p{10cm}|}
\hline
\textbf{Type} & \textbf{Category} & \textbf{Examples} \\
\hline
\multirow{5}{*}{Structured} 
& Static & Age, Gender, Ethnicity, Admission Type, Admission Location, Insurance, Religion \\
\cline{2-3}
& Vitals & Heart Rate, Systolic/Diastolic/Mean Blood Pressure, Respiratory Rate, Temperature, Oxygen Saturation, Capillary Refill Rate \\
\cline{2-3}
& Neurological Scores & Glascow Coma Scale: Eye Opening, Motor Response, Verbal Response, Total \\
\cline{2-3}
& Lab Results & pH, Glucose, Fraction Inspired Oxygen \\
\cline{2-3}
& Physical Measures & Height, Weight \\
& Codes & ICD-9 diagnosis codes, Procedure codes (CPT) \\
\hline
\multirow{2}{*}{Unstructured} 
& Clinical Notes & Discharge summaries, Progress notes, Nursing notes \\
\cline{2-3}
& Conditions & respiratory failure, complications of surgical procedures or medical care, aspiration pneumonitis, etc\\
& Procedures & tracheostomy, respiratory, intubation and mechanical ventilation, gastrostomy etc\\
& Medications & hypnotics, sedatives;  drugs for constipation, other nutrients in atc,adrenergics, inhalants, etc \\

\hline
\end{tabular}
\end{table*}

\begin{table}[ht]
\centering
\caption{ICD-9 Code Categories Used for Comorbidity Flags}
\label{tab:icd_categories}
\begin{tabular}{|l|p{8cm}|}
\hline
\textbf{Category} & \textbf{ICD-9 Prefixes (Examples)} \\
\hline
Explicit Sepsis & 99591, 99592, 78552, 038 \\
Infection & 001--139 (e.g., 001, 008, 038, 041, 079) \\
Organ Dysfunction & 584, 7855, 570, 572, 51881, 7991, 293 \\
Diabetes & 250 \\
Cardiovascular & 390--429 (e.g., 401, 410, 414, 427) \\
Cancer & 140--189 (e.g., 153, 162, 174, 185) \\
Lung Disease & 490--505 (e.g., 491, 493, 496, 500) \\
Dementia & 290, 294 \\
Kidney Dialysis & 585 \\
Liver Disease & 570, 571, 572 \\
Immune Disorder & 279 \\
\hline
\end{tabular}
\end{table}

\subsection{Untructured Data Representation for $M_1$}
\label{M_1:appendix}
In the main paper, we described multiple stages of unstructured text encoder, $M_1$. In this section, we will provide, examples on each step.

\subsubsection{Generating Context}
In this section, for each patient $P$'s visit $v_i$, we utilize information from the MIMIC-III dataset, including ICD codes and physician notes. These are used to extract the patient’s conditions, procedures, and prescribed drugs. Note that not all visits contain physician notes; in such cases, we leave the note field blank. If available, we include the note for the corresponding visit.

For patients with multiple visits, we aggregate information from all prior visits. This means a single entry may contain multiple physician notes concatenated together. However, these notes are often unstructured and may include irrelevant or redundant content, which can lead to context length issues when used as input to large language models (LLMs) during fine-tuning.

\begin{figure*}
    \centering
\includegraphics[width=\linewidth]{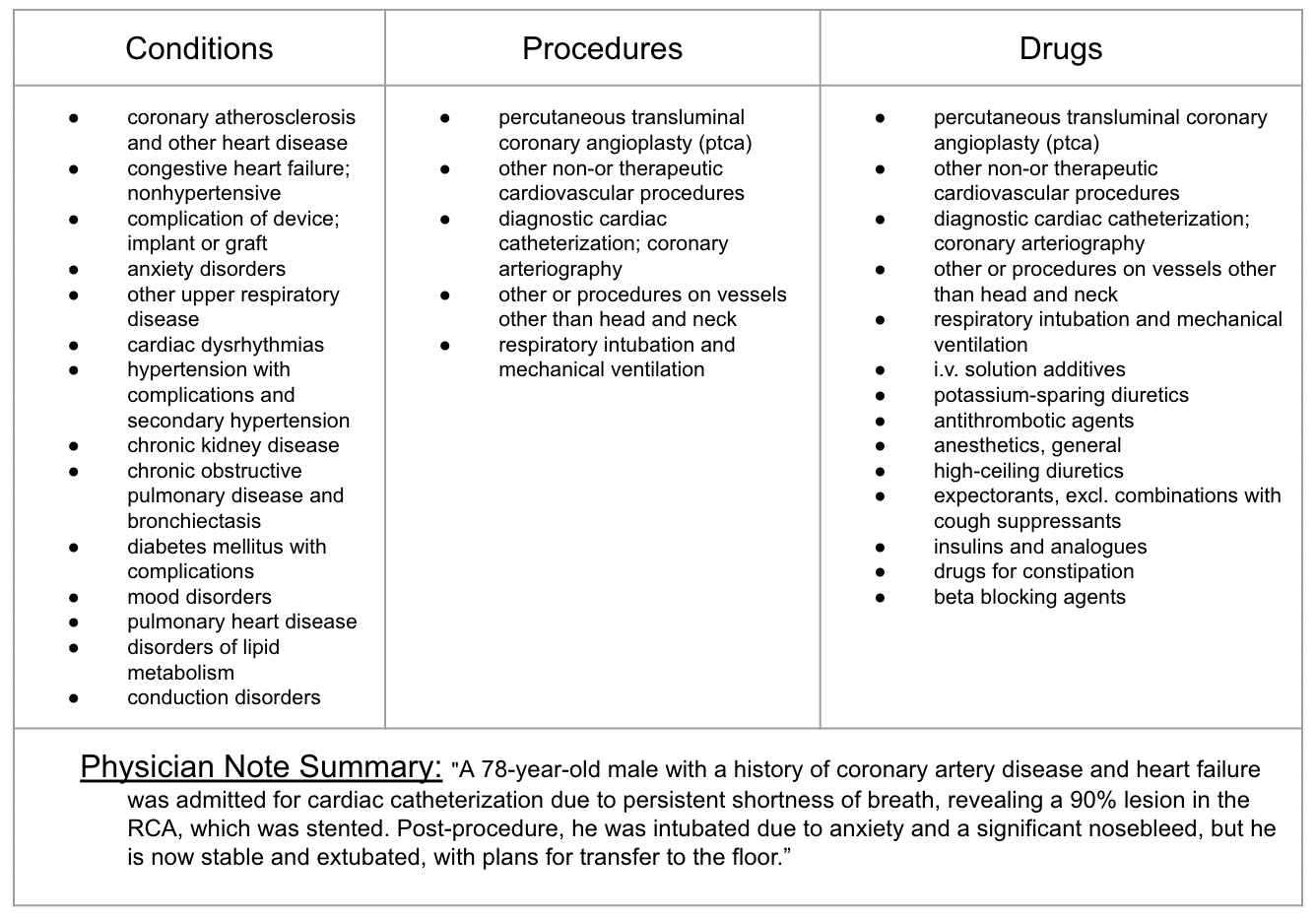}
    \caption{A sample generated context with physician summary for a single Patient Visit, for $M_1$ }
    \label{fig:context}
\end{figure*}

To address this, we employ a separate LLM to summarize the concatenated physician notes, producing a concise and clinically relevant summary for each patient history. Figure \ref{fig:context} demonstrates a sample example of a patient context.

\subsubsection{Knowledge Graph Retrieval}

The first step in our approach involves downloading the latest PubMed dataset and extracting all available abstracts. The goal is to capture general medical knowledge that may support diagnostic reasoning in complex clinical cases.

\begin{figure*}
\includegraphics[width=\linewidth]{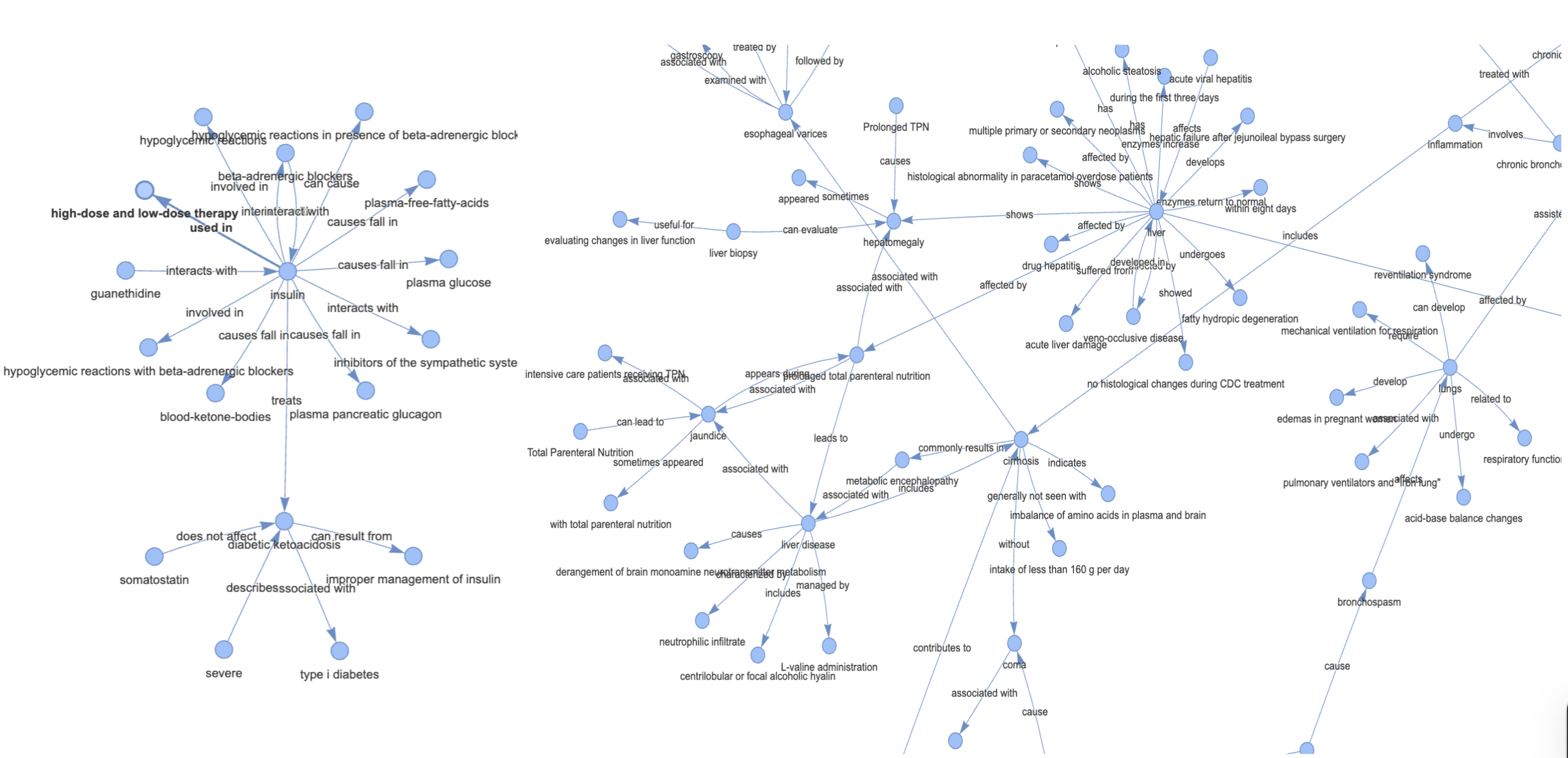}
\caption{A partial snapshot of a knowledge graph built from PubMed data, filtered to include only patient-related concepts.}
\label{fig:KG-appendix}
\end{figure*}

\begin{itemize}
    \item \textbf{Concept Generation}: After constructing the base context from these abstracts, we treat extracted medical conditions, procedures, and other relevant entities as concepts. In addition to the KARE framework, we further enhance concept extraction from patients' physician notes by applying a dedicated NLP model to parse physician notes and extract clinical keywords and concepts directly from unstructured text. These are extracted using a biomedical natural language processing pipeline built on \textit{scispaCy}. Each note is processed with the $en\_core\_sci\_sm$ model to identify entities such as diseases, medications, procedures, symptoms, and conditions. Recognized terms are normalized to lowercase and stored as concepts. For notes where specific biomedical entities are not labeled, relevant noun phrases (longer than three characters) are also included to ensure broader coverage of meaningful medical concepts. These extracted concepts are later combined with structured EHR codes for downstream analysis.
    \item \textbf{KG Construction}: Using these concepts, we perform targeted retrieval from the PubMed abstracts, identifying paragraphs that are semantically aligned with each concept. Once relevant text is retrieved, we employ a large language model (LLM) to extract relational triples (subject, predicate, object), thereby constructing a knowledge graph grounded in biomedical literature. Figure \ref{fig:KG-appendix} shows a partial snapshot of a knowledge graph constructed from PubMed data, filtered to include only patient-related concepts.

    \item \textbf{KG structure}: This PubMed-derived knowledge graph supplements patient-specific data and improves reasoning by providing broader clinical context. The following triples are an example how the relations are extracted, involving \textit{beta blocking agents} from biomedical literature.
\begin{itemize}
    \item (unstable angina pectoris, treated with, \textbf{beta blocking agents})
    \item \textbf{beta blocking agents}, should be avoided in, non responsive patients)
    \item (\textbf{beta blocking agents}, could be a useful measure in, patients with labile arterial hypertension)
    \item (\textbf{beta blocking agents}, could be a useful measure in, atients with vegetative dysregulation)
    \item (\textbf{beta blocking agents}, could be a useful measure in, patients with hyperkinetic heart syndrome)
    \item (bunitrolol, is a type of, \textbf{beta blocking agents})
    \item (\textbf{beta blocking agents}, impact, myocardial lactate extraction)
    \item (\textbf{beta blocking agents}, reduce, arterial NEFA levels)
\end{itemize}

\end{itemize}

\subsubsection{Finding Similar Patients}

Here, the model identifies similar patients by leveraging diagnoses, procedures, and medications, along with unstructured data from physician notes. For each patient visit, KARE extracts key clinical concepts and compares them across the cohort to find patients with overlapping medical histories and treatments. This similarity assessment enables more personalized predictions by grounding the model in relevant patient contexts.

\begin{table}[ht]
\centering
\small
\caption{Example of Similar Patient Visits (both positive and negative) and Clinical Features for readmission in 30 days case}
\label{tab:patient_similarity}
\begin{tabular}{|p{2cm}|p{6cm}|p{6cm}|}
\hline
\textbf{Feature} & \textbf{Patient ID: 12081\_1, Visit 0} & \textbf{Patient ID: 12659\_0, Visit 0} \\
\hline
\textbf{Conditions} & Congestive heart failure, acute cerebrovascular disease, respiratory failure, COPD, cardiac dysrhythmias, coronary atherosclerosis, conduction disorders & Benign neoplasm, heart valve disorders, hepatitis, hypertension, lipid metabolism disorders \\
\hline
\textbf{Procedures} & Pacemaker insertion/revision, blood transfusion, cardiac rhythm conversion & Heart procedures, extracorporeal circulation, blood transfusion \\
\hline
\textbf{Medications} & ACE inhibitors, adrenergics (inhalants), corticosteroids, lipid modifiers, cardiac stimulants, potassium supplements, antithrombotics, opioid analgesics & Antiinfectives, ACE inhibitors, beta blockers, calcium channel blockers, diuretics, analgesics, anxiolytics \\
\hline
\textbf{Physician Notes} & Continued respiratory failure, CHF, COPD; new lung cancer, heart valve disorders, diagnostic bronchoscopy, new medications & Not available \\
\hline
\textbf{Label} & 1 & 0\\
\hline
\end{tabular}
\end{table}

\begin{figure}
    \includegraphics[width=1\linewidth]{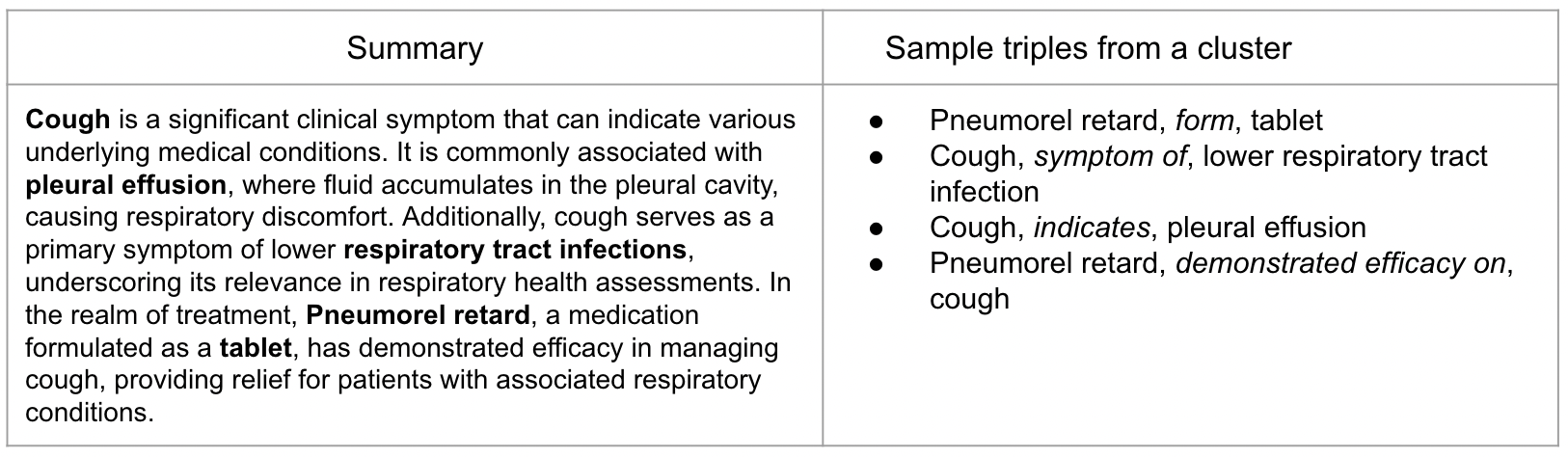}
    \caption{A cluster from the Knowledge Graph generated using the Leiden clustering algorithm, along with the corresponding LLM-generated summary of its triples.}
    \label{fig:placeholder}
\end{figure}


\subsubsection{Reasoning Module}
A critical component of our framework is the reasoning module, which enables the model to go beyond surface-level pattern matching and perform clinically informed inference. This module integrates the  patient data—including summarized clinical notes, and the retrieved biomedical knowledge graph—to generate logical, interpretable explanations supporting the risk predictions.

Specifically, the reasoning module processes extracted clinical concepts and relationships from patient history and external knowledge to construct chains of evidence, mimicking clinician decision-making steps. This approach helps the model distinguish relevant signals from noise in complex and heterogeneous data.

In Figure~\ref{fig:full_task}, the reasoning module is illustrated as a dedicated section within the overall prompt structure. It explicitly asks the LLM to produce detailed, medically grounded reasoning before outputting the final risk prediction, which improves both accuracy and interpretability. Ablation studies (Table~\ref{tab:ablation}) demonstrate that removing this reasoning component significantly degrades predictive performance, underscoring its essential role.

\subsection{Full Example Prompt}
\begin{figure*}[htbp]
    \centering
\includegraphics[width=1.05\linewidth, height=13cm]{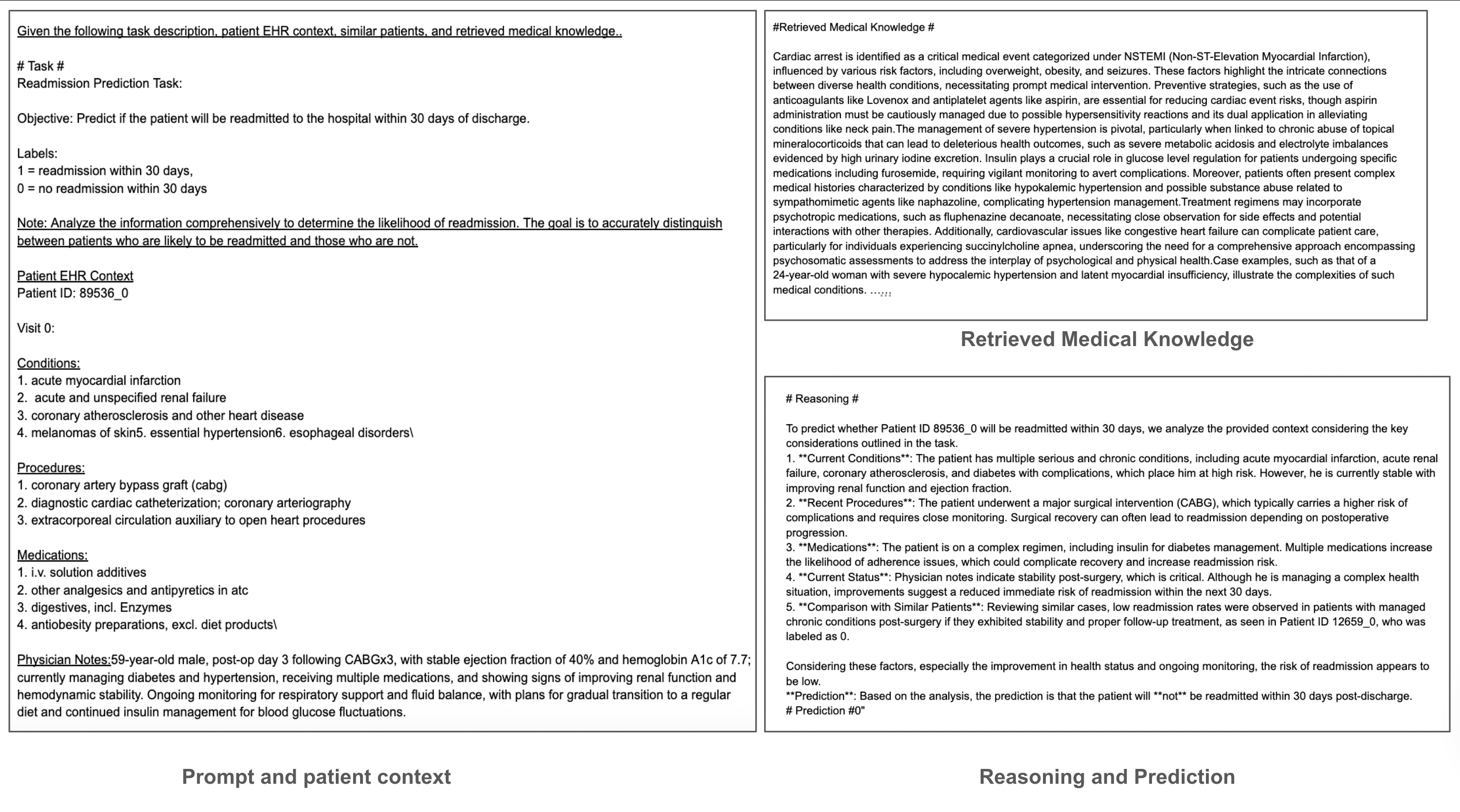}
    \caption{Example of the overall prompt structure provided to the LLM during fine-tuning, including task description, patient EHR context, retrieved medical knowledge, and expected reasoning and prediction outputs.}
    \label{fig:full_task}
\end{figure*}

Figure~\ref{fig:full_task} illustrates the complete prompt structure used to fine-tune the LLM. The prompt includes \textbf{(1)} a clear task description; \textbf{(2)} the target patient’s electronic health record (EHR) context combining structured features and summarized clinical notes; \textbf{(3)} relevant external biomedical knowledge retrieved from PubMed abstracts represented as a knowledge graph; \textbf{(4)} examples of similar and dissimilar patient cases for comparative context (We show this in Table \ref{tab:patient_similarity}; and \textbf{(5)} reasoning behind the prediction, generated from an LLM. 

The prompt concludes with an explicit request for medically grounded reasoning followed by the final risk prediction. This structured input design facilitates interpretable and accurate model outputs.

\subsection{Structured Data Representation for $M_2$}

We preprocess raw patient records by extracting structured data from MIMIC‑III and organizing it into well-defined prediction datasets. We align each patient to an admission based on predefined cohort criteria, for in‑hospital mortality prediction, we use early admission data, typically features collected within the first 24 hours, to build static representations. These include demographic attributes (such as age, gender, and admission type), physiological measurements (vital signs like heart rate or blood pressure), lab test results, and interventions. All continuous variables are normalized and discretized to create time‑series bins or summary statistics. For readmission prediction, we rely on discharge time data or early records filtered using exclusion criteria (e.g., patients under 18 or with missing data), and construct 30‑day readmission labels by linking consecutive admissions. 

To handle missing values, we apply simple imputation strategies to ensure a consistent feature space across all patients. Specifically, missing entries in time-series data are filled using forward-fill and backward-fill methods where possible. For variables still missing after temporal imputation, we use either zero imputation (after normalization) or impute with population-level statistics such as the mean or median, depending on the variable type. This approach ensures that the downstream models receive complete input tensors without discarding any patient records due to sparsity, which is critical in clinical data settings where missingness is common and often informative.

\noindent
\textbf{Incorporating Patient Demographics and Admission Details}
This step involves enriching the dataset with key patient demographics and hospital admission attributes. Then we merge external data tables containing information such as gender, ethnicity, admission type (emergency, elective), insurance, and language. Categorical variables are converted into numeric codes for easier model consumption, while missing values are handled with placeholder categories to maintain data integrity. This results in a structured feature set that captures essential background and administrative details for each patient episode.

\noindent
\textbf{Adding Clinical Diagnoses, Procedures, and Medication History}
Next, the dataset is augmented by integrating clinical history through diagnosis codes, procedure codes, and prescribed medications linked to each hospital admission. For instance, all ICD diagnosis codes recorded during a hospital stay are collected into a list, allowing the creation of binary flags that indicate the presence of major conditions like diabetes or cardiovascular disease. Additionally, common code prefixes are extracted to generate features representing groups of related diagnoses or procedures. Similarly, medication records are summarized into key drug groups. This approach transforms complex clinical histories into a numerical format that models can use to assess patient risk.

\noindent
\textbf{Embedding Clinical Reasoning and Merging Predictions}
The final enrichment involves processing free-text clinical reasoning or explanations associated with predictions. These textual rationales are tokenized and converted into numerical vector embeddings that capture their semantic meaning, for example, representing the sentence “patient shows signs of infection” as a 100-dimensional vector. These embeddings are paired with binary prediction labels and merged back into the main dataset by matching patient episodes. Missing values are filled with neutral defaults to ensure consistent data structure. The combined result is a dataset where each patient episode has both predictive labels and rich, encoded reasoning features to support improved downstream modeling or interpretation.

\section{Additional results}

\noindent
\textbf{Using Anomaly Detection Method for Readmission in 30 days}:
Given the highly imbalanced nature of the 30-day readmission data, where only approximately $4\%$ of hospital visits result in readmission, we explored whether this task could be framed as an anomaly detection problem. Specifically, we investigated whether the minority class (readmitted patients) exhibits anomalous patterns relative to the majority class. 

To evaluate this, we applied unsupervised anomaly detection methods, such as the Isolation Forest algorithm, to identify readmissions as statistical outliers based on the available clinical features. Table~\ref{tab:dataset_distribution} shows the distribution of positive and negative samples in the training and test sets, along with the number of anomalies detected by the model.

Despite the appeal of this approach, anomaly detection methods did not perform well for this task. The results indicate that the readmitted patients do not form distinct statistical outliers in the feature space, making it difficult for unsupervised models to reliably separate them from the majority class. This suggests that the features available do not naturally differentiate readmissions as anomalous events, and that supervised learning approaches are more appropriate for this problem.

\begin{table}[h!]
\centering
\resizebox{0.8\linewidth}{!}{%
\begin{tabular}{lccccc}
\hline
\textbf{} & \textbf{Positive} & \textbf{Negative}  & \textbf{Anomalies Detected} & \textbf{Actual positive} \\
\hline
Train Set& 718   & 17185 & 717 & 33 \\
Test Set& 123   & 3113  & 130 & 3 \\
\hline
\end{tabular}
}
\caption{Distribution of positive and negative samples across train, test, and anomaly detection settings using Isolation Forest method}
\label{tab:dataset_distribution}
\end{table}

\begin{table}[h!]
\centering
\caption{Inference Results from Fine-Tuned LLM for Mortality Prediction (Part 1)
}
\begin{small}
\begin{tabular}{|p{17cm}|}
\hline
\textbf{Patient ID: 9782\_1} \newline
\#\#\# \textbf{Step 1: Analyzing Conditions}

1. **Septicemia (except in labor)** - This is a severe condition that indicates systemic infection, which can lead to sepsis and high mortality risk. \\
2. **Shock** - Indicates a critical state, often associated with poor outcomes. \\
3. **Pneumonia** - This can be serious, especially in the context of septicemia and shock, as it can complicate respiratory function. \\
4. **Schizophrenia and other psychotic disorders** - While this is a mental health condition, it does not directly impact mortality. \\
5. **Deficiency and other anemia** - This can complicate other conditions, especially in the context of infection or respiratory issues. \\
6. **Thyroid disorders** - Generally less critical unless there are acute complications. \\

\#\#\# \textbf{Step 2: Analyzing Procedures}

1. **Other vascular catheterization; not heart** - This is an invasive procedure that can carry risks, especially in the context of the patient's other conditions. \\
2. **No major surgeries or life-sustaining procedures** - The absence of life-sustaining interventions suggests that the patient may not be in immediate critical condition. \\

\#\#\# \textbf{Step 3: Analyzing Medications}

1. **Antibiotics (quinolone antibacterials, other antibacterials)** - Indicate the treatment of infection, which is critical given the septicemia and pneumonia. \\
2. **Corticosteroids for systemic use** - Can be used in severe conditions but also carry risks, especially in the presence of infections. \\
3. **Opioid analgesics** - Often used for pain management, which may indicate severe underlying conditions. \\
4. **Cardiac stimulants** - Suggests potential heart issues. \\
5. **High-risk medications** - The use of multiple medications indicates a complex health status, which increases the risk of adverse effects and complications. \\

\#\#\# \textbf{Step 4: Comparing with Similar Patients}

1. **Patient ID 15209\_0** had conditions such as septicemia, respiratory failure, and shock, leading to a label of 0 (survival). This patient had similar severe conditions but also underwent extensive invasive procedures and had more acute conditions. \\
2. **Patient ID 28061\_0** had coronary atherosclerosis and cardiac issues leading to a label of 1 (mortality). This patient had significant cardiovascular issues and underwent multiple invasive procedures. \\

\#\#\# \textbf{Step 5: Integrating Information} \\
- The combination of septicemia, shock, and pneumonia in Patient ID 9782\_1 is extremely concerning, especially considering the lack of major surgical interventions. The presence of multiple comorbidities (anemia, schizophrenia, and thyroid disorders) complicates the overall picture. \\
- The complexity and severity of the patient's conditions (particularly septicemia and shock) indicate a high risk of mortality, but the absence of life-sustaining procedures suggests some resilience. \\

\#\#\# \textbf{Conclusion} \\
Given the severity and complexity of Patient ID 9782\_1’s conditions, the presence of multiple high-risk medications, and the lack of immediate life-threatening interventions, the prediction leans towards survival (label 0). However, the critical nature of the patient's conditions suggests that the risk of mortality is significant, and a careful monitoring plan is essential for managing the patient's health. \\

\#\#\# \textbf{Final Prediction} \\
- **Prediction**: 0 (survival) \\
\hline
\end{tabular}
\end{small}
\label{tab:inference_table2_part1}
\end{table}

\begin{table}[h!]
\centering
\caption{Inference Results from Fine-Tuned LLM for Mortality Prediction (Part 2)}
\begin{small}  
\begin{tabular}{|p{17cm}|}
\hline
\textbf{Patient ID: 85258\_1} \newline
\#\#\# \textbf{Step 1: Analyze Patient Conditions} \\
1. **Cancer of Esophagus**: This is a critical condition that is often associated with high mortality risk, particularly when combined with other severe conditions. \\
2. **Secondary Malignancies**: The presence of secondary malignancies indicates that the patient has advanced cancer, further increasing the risk of mortality. \\
3. **Pneumonia**: This is a severe infection that can lead to respiratory failure, especially in a patient with compromised respiratory function due to cancer. \\
4. **Respiratory Failure**: The patient has been diagnosed with respiratory failure, indicating a critical state that significantly increases the risk of mortality. \\
5. **Septicemia**: This is a serious systemic infection that can lead to multi-organ failure, which is a key indicator of mortality. \\
6. **Deficiency Anemia**: This can complicate the patient's overall health, particularly in the context of cancer and infection. \\
7. **Other Comorbidities**: The patient has several other comorbidities, including disorders of teeth and jaw, white blood cell diseases, and fluid and electrolyte disorders, which compound the risk of mortality. \\

\#\#\# \textbf{Step 2: Evaluate Procedures} \\
1. **Tracheostomy**: This is a significant surgical intervention that indicates severe respiratory compromise and is often performed in critically ill patients. \\
2. **Mechanical Ventilation**: The patient is on mechanical ventilation, which is a life-sustaining measure typically reserved for critically ill patients. \\
3. **Blood Transfusions**: The need for multiple blood transfusions suggests significant anemia or other severe conditions that require urgent intervention. \\

\#\#\# \textbf{Step 3: Assess Medications} \\
1. **Broad Spectrum Antibiotics**: The patient is receiving broad-spectrum antibiotics, which are often used in the presence of sepsis or severe infections. \\
2. **Corticosteroids**: The use of corticosteroids suggests a significant inflammatory response or severe underlying condition. \\
3. **Antipsychotics and Antidepressants**: These medications suggest a complex mental health component, which can complicate treatment and recovery. \\

\#\#\# \textbf{Step 4: Compare with Similar Patients} \\
1. **Patient ID: 28061\_0**: This patient had a history of coronary artery disease and complications from medical care, leading to a mortality label of 1. The presence of critical cardiac issues and multiple comorbidities mirrors the complexity of Patient ID: 85258\_1. \\
2. **Patient ID: 14297\_0**: This patient had several chronic conditions but ultimately had a label of 0, indicating that not all patients with complex conditions lead to mortality. The complexity of Patient ID: 85258\_1, however, suggests a much higher risk. \\

\#\#\# \textbf{Step 5: Synthesize Findings} \\
- The combination of advanced cancer, severe infections (septicemia), respiratory failure, and invasive procedures (tracheostomy and mechanical ventilation) strongly indicates that the patient is in a critical state. \\
- The presence of multiple comorbidities and the complexity of the medications suggest that the patient is in a high-risk category for mortality. \\

\#\#\# \textbf{Conclusion} \\
Given the severity of the conditions, the invasive nature of procedures, and the complex medication regimen, the patient is at an extremely high risk of mortality. Therefore, the prediction for Patient ID: 85258\_1 is: \\

**Prediction**: 1 (mortality) \\
\hline
\end{tabular}
\end{small}
\label{tab:inference_table2_part2}
\end{table}

\section{Related Works}

\noindent \textbf{Clinical Prediction using Statistical and Machine Learning Methods}: Structured data collected from electronic health records (EHRs) contain valuable information about patients and can be instrumental in capturing patient states and predicting medical outcomes (e.g., mortality)~\cite{jiangreasoning,qiu2025deep,yang2023transformehr}. To leverage these data, traditional machine learning models—such as linear regression~\cite{brisimi2018predicting}, decision trees~\cite{herazo2021decision}, and Bayesian approaches~\cite{bellot2019bayesian}—use structured features as input to model the complex relationships between patient features and clinical outcomes. More recently, deep learning–based frameworks have also been explored to capture these complex dynamics and have shown great potential~\cite{cui2025identifying,arsalan2025enhancing,yu2024smart,wu2023medlink,zhu2024emerge}. However, deep learning-based approaches remain largely restricted to structured data and rely heavily on statistical correlations between input features and outcome labels. Deep learning-based approaches also often lack interpretability grounded in explicit clinical knowledge and tend to be highly specific to the training dataset, limiting their generalizability to other clinical settings~\cite{jiangreasoning,jiang2024graphcare}.

\noindent \textbf{NLP and Language Models}: Unstructured clinical texts (e.g., physician notes, radiology reports, and nursing progress notes)  contain rich information about patients ~\cite{johnson2016mimic,liu2024time,jiang2024graphcare}. Textual documentation includes clinicians’ reasoning and management plans that are not contained elsewhere in the EHR and which may provide direct and high-value insights. To harness this information, various natural language processing (NLP) methods have been proposed for predictive modeling~\cite{mullenbach2018explainable,lipton2016learning,yao2018clinical}. Recently, large language models (LLMs) have demonstrated significant progress across a range of tasks, and researchers have begun exploring their potential for processing unstructured clinical text and supporting clinical reasoning~\cite{liu2024time,niu2024ehr,xu2024ramehr}. However, one of the major challenges in applying LLMs to healthcare is the risk of errors and hallucinations~\cite{cui2025llms,shi2024ehragent}. To address this, some works incorporate knowledge graphs to explicitly embed clinical knowledge and mitigate LLM inaccuracies~\cite{soman2024biomedical,jiang2024graphcare}. Despite these advances, integrating both structured EHR data and unstructured clinical text in a unified framework remains a challenging problem~\cite{zhang2021benchmarking}. Existing approaches—such as prompt engineering~\cite{hu2024improving}, RAG~\cite{zhu2024emerge}, and data fusion~\cite{cui2024multimodal,thao2024medfuse}—tend to be ad hoc and lack a systematic method for joint modeling.


\end{document}